\pdfoutput=1

\documentclass[11pt]{article}

\usepackage[final]{acl}

\usepackage{times}
\usepackage{latexsym}

\usepackage[T1]{fontenc}

\usepackage[utf8]{inputenc}

\usepackage{microtype}

\usepackage{inconsolata}

\usepackage{graphicx}

\usepackage{soul} 

\usepackage{graphicx}
\usepackage{caption}
\usepackage{subfigure}
\usepackage{algorithm}
\usepackage{algorithmic}
\usepackage{multirow}
\usepackage{pifont}
\usepackage{bbding}
\usepackage{tabularx} 
\usepackage{colortbl}
\usepackage{wrapfig}
\usepackage{booktabs}

\usepackage[most]{tcolorbox}
\newtcolorbox{response}[1][]{
  colback=gray!5,
  colframe=black,
  fonttitle=\bfseries,
  coltitle=black,
  }

\sethlcolor{yellow}

\title{FedDQC: Data Quality Control in Federated Instruction-tuning of \\Large Language Models}

\author{
 \textbf{Yaxin Du\textsuperscript{1}},
 \textbf{Rui Ye\textsuperscript{1,3}},
 \textbf{Fengting Yuchi\textsuperscript{1}},
 \textbf{Wanru Zhao\textsuperscript{2}},
\\
 \textbf{Jingjing Qu\textsuperscript{3}},
 \textbf{Yanfeng Wang\textsuperscript{1}},
 \textbf{Siheng Chen\textsuperscript{1*}},
\\
 \textsuperscript{1}Shanghai Jiao Tong University,
 \textsuperscript{2}University of Cambridge,
 \textsuperscript{3}Shanghai AI Laboratory
\\
 \small{
   \textbf{Correspondence:} \href{sihengchen@sjtu.edu.cn}{sihengchen@sjtu.edu.cn}
 }
}

\begin{document}
\maketitle
\begin{abstract}
Federated Learning (FL) enables privacy-preserving collaborative instruction tuning of large language models (LLMs) by leveraging massively distributed data. However, the decentralized nature of FL exacerbates data quality challenges, as local clients lack global visibility to filter noisy or low-quality samples before training. To resolve this issue, we propose FedDQC, a novel federated instruction tuning framework with dynamic data quality control. Our approach introduces two key innovations. First, we propose instruction-response alignment (IRA)—an efficient client-side metric for quality evaluation requiring only low-cost inference. We validate that higher-IRA data corresponds to more relevant and easier-to-learn question-answer pairs. Second, mirroring the human easy-to-hard knowledge acquisition process, we design a quality-aware hierarchical FL training framework, where the LLM is progressively fine-tuned from high- to low-IRA data in a collaborative manner. The framework also supports adaptive data quality assessment at each hierarchy, enabling dynamic adjustments throughout the training process. Extensive experiments on synthetic and real-world datasets show that our method significantly improves LLM performance on mixed-quality data in FL.
\end{abstract}

\section{Introduction}
\vspace{-0.1cm}
For large language models (LLMs) training~\cite{roumeliotis2023chatgpt, chowdhery2023palm, touvron2023llama, jiang2023mistral}, both the quantity and quality of the training data significantly impact their performance~\cite{zhao2023survey,minaee2024large}. The scaling law suggests that more training data can lead to more powerful LLMs~\cite{kaplan2020scaling}. However, in specific domains such as healthcare~\cite{llm_medicine} and finance~\cite{wu2023bloomberggpt}, privacy concerns~\cite{albrecht2016gdpr} prevent the aggregation of large-scale datasets, making it challenging to expand the dataset scale. Federated Learning (FL)~\cite{fedavg}, as an emerging distributed training approach, preserves privacy by allowing multiple clients to train a unified model collaboratively without sharing their data. This enables dataset scaling while ensuring data privacy~\cite{chen2023federated, openfedllm,kuang2024federatedscope,fan2023fate,ye2024leveraging}.

\begin{figure}[t]
    \includegraphics[scale=0.42]{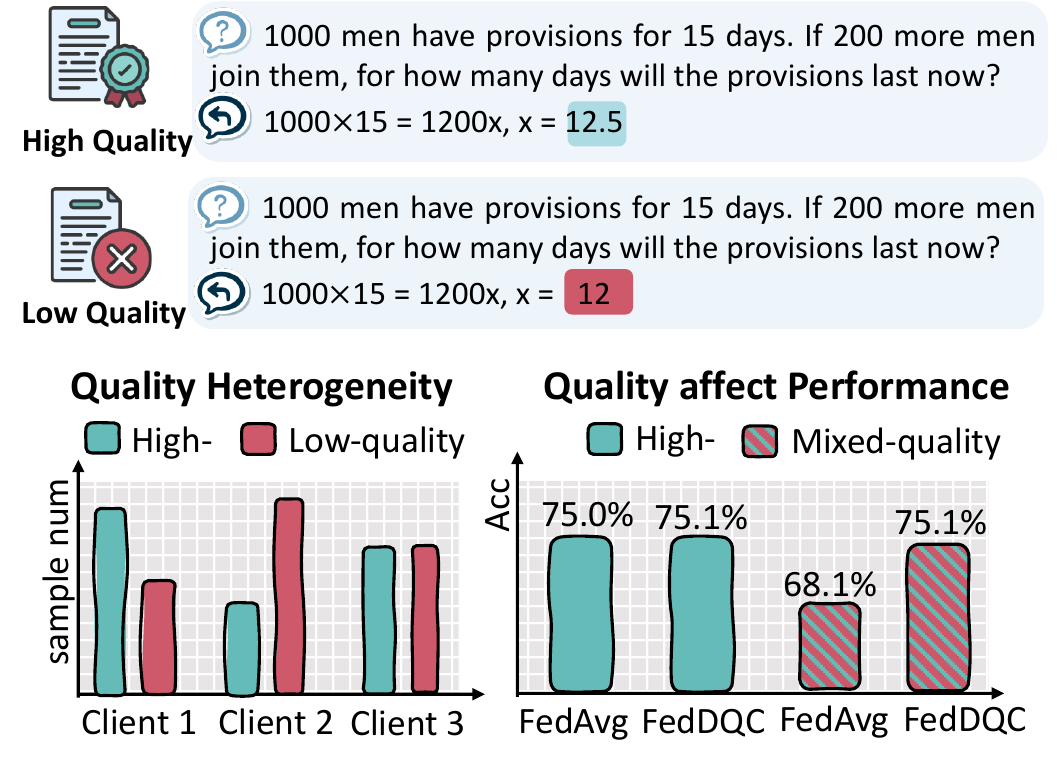}
    \vspace{-0.4cm}
    \label{fig:fig1}
    \caption{Top figure is an example of low-quality data and high-quality data. The left figure shows federated quality heterogeneity. The right figure shows how data quality affects federated training performance and FedDQC eliminates low-quality data effects.}
    \vspace{-0.8cm}
\end{figure}

While FL addresses the data quantity issue by incorporating more local clients, it brings more data quality issues~\cite{shaheen2022applications}. In FL, training data for each client are collected from various sources locally, making it difficult to detect low-quality data or noises in local datasets. Such vulnerabilities adversely affect general model training. Although numerous methods~\cite{wu2023self, liu2023makes, cao2023instruction, chen2023maybe, zhou2024lima} are proposed for data quality control in LLM instruction tuning, their designs typically require access to the entire training data, making them impractical for FL scenarios. Therefore, in this work, we aim to bridge this gap and address the under-explored issue of federated data quality control in instruction-tuning LLM tasks.

Existing data quality control methods focus on designing data quality evaluation metrics to quantify data quality. They can be broadly categorized into two types. The first category consists of heuristic-based methods specifically designed for instruction-tuning tasks.~\cite{wang2024survey} These methods quantify the quality of instruction-response pairs~\cite{IFD, du2023mods, cao2023instruction} by quantity of information. However, they rely on the assumption that all data are clean and reliable making them difficult to deal with noises and errors in dataset. The second category is traditional data attribution methods~\cite{kwon2023datainf, ghorbani2019data, ilyas2022datamodels}, which require re-training and evaluation on the whole dataset. However, in FL each client only has limited computation resources for re-training and does not allow access to the local dataset, making these methods impractical for FL.

To fill this gap, we propose FedDQC (Federated Data Quality Control), a novel FL framework with data quality control for LLM instruction tuning. First, we propose an efficient and privacy-preserving data quality scoring metric: IRA (Instruction-Response Alignment), which could be computed on the client side with minimum cost. This metric evaluates the data quality by estimating the mutual information between the instruction and response on LLM. Specifically, it calculates the response inference loss difference between given instruction and without instruction. This approach fully leverages the knowledge embedded in the pre-trained LLM and eliminates the impact of response format inconsistency with pre-trained data. In the context of instruction tuning, higher alignment indicates that the instruction and response are better matched, enabling the model to learn how to answer questions more easily.

Based on the proposed IRA scores, we propose an FL training framework fully leveraging data quality evaluation to handle data quality issues. The key idea behind FedDQC is a combination of hierarchical training and adaptive scoring during training. Specifically, it consists of two stages: the scoring stage and the hierarchical training stage. In the scoring stage, each client independently computes IRA scores for local data samples using the current global model through localized inference without requiring external data access. All data is then re-ordered based on IRA scores, and high-quality data is selected and partitioned into hierarchical subsets for subsequent training. in the hierarchical training stage, the model initially trains on high-quality, easily learnable samples (higher IRA scores) and gradually transitions to more complex data (lower IRA scores). This staged approach reduces interference from challenging samples in early training phases, thereby enhancing learning efficiency. These two stages iterate until the final hierarchy, effectively enabling adaptively scoring during training. This progressive knowledge integration mechanism allows the model to incorporate more challenging data as its capacity improves. This leads to enhanced model robustness and performance, particularly when dealing with noisy or heterogeneous data.

Our experiments demonstrate that FedDQC not only outperforms all baseline models in both IID (independent and identically distributed) and non-IID settings on four synthetic datasets but also shows effectiveness on the real-world federated dataset, Fed-WildChat~\cite{fedllm-bench}. As for computation, we show that the scoring metric IRA consumes only 1\% training time for data quality evaluation, making it computation-efficient and scalable for larger datasets.

\begin{figure*}[tb]
    \centering
    \vspace{-1cm}
    \includegraphics[scale=0.42]{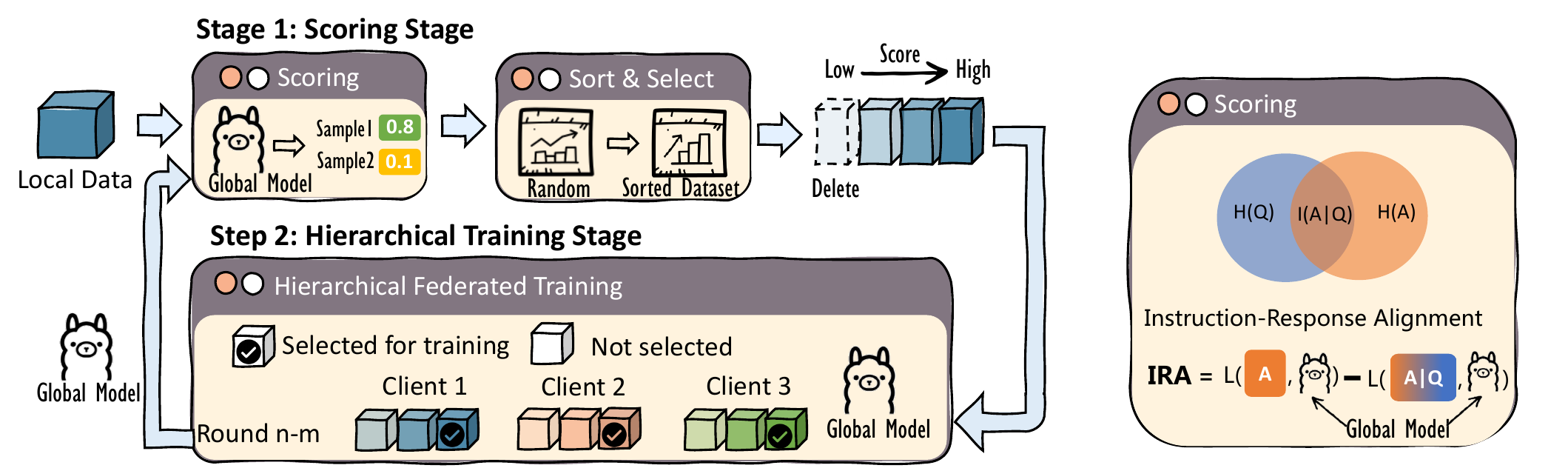}
    \vspace{-0.3cm}
    \caption{Overview of FedDQC, which iterates in two stages: (1) Scoring stage: utilize IRA and global model to evaluate data quality; (2) Hierarchical training: progressively fine-tuned from high-IRA to low-IRA data, mirroring the easy-to-hard learning process; (3) Scoring stage and hierarchical training stage iterates to the last hierarchy.}
    \vspace{-0.55cm}
    \label{fig:main}
\end{figure*}

\vspace{-0.1cm}
\section{Related work}
\vspace{-0.1cm}
\subsection{Federated Learning}
\vspace{-0.1cm}
Federated Learning~\cite{kairouz2021advances,fedavg,li2020federated} has emerged as a powerful method for privacy-preserving collaborative training, allowing multiple clients to jointly train a global model without sharing raw data, coordinated by a central server. Existing research on data quality in FL primarily focused on the classification tasks, with noisy label issues.~\cite{li2021sample} We classify related data quality control works from three levels: client, model and sample level. At the client level, efforts have concentrated on identifying malicious clients~\cite{jiang2023data, yang2021client} through feature~\cite{yang2022robust} or model weight clustering~\cite{wang2022fednoil}.
While at the sample level, studies have typically focused on label correction strategies~\cite{xu2022fedcorr} or confidence-based sample reweighting~\cite{fang2022robust}. At the model level, approaches like distillation~\cite{wang2024federated} or modifying the loss function~\cite{wu2023learning} aimed to increase robustness against noisy labels. However, these methods do not effectively address the unique challenges of federated LLM training, the generation task. This highlights the gap in current approaches and underscores the need for specialized solutions tailored to generative tasks in FL.

\vspace{-0.1cm}
\subsection{Data quality control}
\vspace{-0.1cm}

Data quality control is complex and a throughout problem in machine learning~\cite{zha2023data}. To solve the task for this work, we split the related work into two lines: the traditional data attribution with its adaptation to LLM setting, and current data selection work for LLM.

\vspace{-0.2cm}
\paragraph{Data attribution}
Traditional data attribution methods, used to explain model predictions by identifying influential training examples, are generally categorized into retraining-based and gradient-based techniques. ~\cite{hammoudeh2024training} Retraining-based approaches, such as leave-one-out~\cite{ling1984residuals}, Shapley value~\cite{ghorbani2019data}, and Datamodels~\cite{ilyas2022datamodels}, estimate the effect of data points by repeatedly retraining the model on different subsets of data. These data attribution approaches are post-hoc and computationally costly, making them unsuitable for LLM setting.
Gradient-based approaches, like represented point selection~\cite{yeh2018representer}, TracIn~\cite{pruthi2020estimating}, and influence functions~\cite{koh2017understanding}, estimate training data's impact through parameter sensitivity.  
Recent studies have developed more efficient adaptations of this gradient-based method for generative tasks~\cite{guo2020fastif} and LLM settings, streamlining data selection processes such as pre-training~\cite{park2023trak} and instruction-tuning in transfer learning scenarios~\cite{xia2024less}.
Despite these advancements in reducing computational complexity through approximations, computing these methods for LLM data selection is still costly due to the increasing size of large model and data volumes.

\vspace{-0.2cm}
\paragraph{Data selection for LLMs}
Current data selection works for LLM instruction-tuning are heuristic and aimed at core set selection. They either depend on a powerful external model for scoring or require iterative training or selection. External model-based scoring techniques, such as AlpaGasus~\cite{chen2023alpagasus}, DEITA~\cite{liu2023makes} and INSTAG~\cite{lu2023instag} prompt ChatGPT~\cite{roumeliotis2023chatgpt} for various dimension of data quality scoring. While effective, these methods are costly and compromise privacy by requiring direct data sharing. This is particularly problematic in privacy-sensitive settings.
Other methods that comply with privacy constraints still require large computation and are not well-suited for local dataset management essential in FL environments.
For instance, IFD~\cite{IFD} and MoDS~\cite{du2023mods} require a computationally intensive initial training stage that may involve low-quality data. 
Similarly, InstructionMining~\cite{cao2023instruction} despite utilizing innovative statistical regression to fit quality influence factors with performance, is dataset-specific and requires retraining. 
Additionally, approaches like SelectIT~\cite{liu2024selectit} and NUGGETS~\cite{nuggets} utilize in-context learning but highly depend on the predefined task set, which is sometimes applicable for FL.
These challenges underscore the need for a new, locally implementable, efficient scoring method that preserves privacy and reduces computational overhead.

\vspace{-0.3cm}
\section{Problem formulation}
\vspace{-0.2cm}
\subsection{Preliminary: Federated Learning}
We consider there are $N$ clients participating in FL to collaboratively train a model $\theta$. Each client holds a dataset $\mathcal{D}_n$ and optimizes its local model via a loss function $L(\cdot)$. The goal of FL is to find the optimal global model $\theta^*$ that minimizes the aggregated loss of all clients. Mathematically, the global objective of FL is:
\begin{equation*}
    \theta^* = \mathop{\arg \min}_{\theta}\sum^N_{n=1} \frac{w_n}{|\mathcal{D}_n|} \underset{{x\in\mathcal{D}_n}}{\sum} L(x, \theta)
\end{equation*}

\vspace{-0.4cm}

\noindent where $w_n=\frac{|\mathcal{D}_n|}{\sum_{i=1}^N|\mathcal{D}_i|}$ represents the weight assigned to client,  $|\mathcal{D}_n|$ is dataset size of $\mathcal{D}_n$.

In the basic FedAvg, each training round $r$ proceeds as follows: 1) Sever broadcasts the global model $\theta^r$ to clients; 2) Each client $n$ performs local model training using $t$ SGD steps to obtain a trained model denoted by $\theta^{r,t}$; 3) Clients upload the locally trained models $\theta^{r,t}$ to the server and the server updates the global model based on the aggregated local model: 
$\theta^{r+1}=\sum_{n=1}^N w_n \theta_n^{r,t}$.

\vspace{-0.2cm}
\subsection{Federated Instruction Tuning}
\vspace{-0.1cm}
\label{Sec:IT_def}
In federated instruction tuning, each client holds a dataset where each sample
is a pair: (question, answer). For client $n$, the dataset is denoted as $\mathcal{D}_n=\{(q^i, a^i)|i=1,2,\dots,|\mathcal{D}_n|\}$, where $q^i$ and $a^i$ denote the $i$-th instruction and answer. The instruction tuning training loss for the $i$-th sample is formulated as $L((a^i, q^i),\theta)=-\sum_{j=1}^{l_i} \log p(a^i_j|q_i\oplus a^i_{<j};\theta)$, where $\oplus$ is the concatenation operator, $l_i$ is the token length of output $a^i$ and $a^i_{<j}$ denotes the tokens before index $j$.

\vspace{-0.1cm}
\section{Methodology}
\vspace{-0.1cm}
This section presents the two-stage FedDQC framework: the scoring stage and the training stage. Firstly, Section~\ref{sec:overview} gives an overview of FedDQC. Then Section~\ref{sec:score} and Section~\ref{sec:train} introduce the scoring  and training stage respectively.

\vspace{-0.2cm}
\subsection{Overview}
\vspace{-0.1cm}

\label{sec:overview}

FedDQC operates through an iterative two-stage process: the scoring stage and the training stage. These stages alternate in a continuous cycle, allowing the model to select and progressively learn from high-quality data.

\noindent\textbf{Scoring Stage:} At the beginning of each hierarchy, clients assess the quality of their local data using the IRA metric, which evaluates the alignment between instructions and responses. Based on these scores, clients sort and filter their data, retaining only high-quality samples for federated training.

\noindent\textbf{Training Stage:} Client partition the filtered high-quality local data into several subsets based on sorted sequence, with each subset with equal size. In each hierarchy training, clients only choose the highest-scored subset to participate in federated training in this hierarchy. 

Please refer to Fig.~\ref{fig:main} and an algorithmic summary in Algorithm~\ref{alg:algorithm}.

\vspace{-0.2cm}
\subsection{Scoring Stage: Data Quality Assessment}
\vspace{-0.1cm}
\label{sec:score}

FedDQC controls data quality in FL by locally assessing data under privacy and computation constraints, allowing clients to select and sort data based on quality.

\begin{table*}[t]
\begin{center}
\caption{Performance comparisons on real and synthetic datasets in both IID and NIID settings show that FedDQC outperforms all methods and even surpasses full clean data training. The best performance for each data quality control method is bolded.}
\vspace{-10pt}
\scalebox{0.62}{
\begin{tabular}{l|c|cccccccc}
\hline
\multicolumn{1}{c|}{}                           & \textbf{Real Dataset}                  & \multicolumn{8}{c}{\textbf{Sythetic Dataset}}                                                                                                                                                                                                                                                                              \\ \hline
\multicolumn{1}{c|}{\multirow{3}{*}{\textbf{}}} & \multirow{2}{*}{\textbf{Fed-WildChat}} & \multicolumn{4}{c|}{\textbf{IID}}                                                                                                                                      & \multicolumn{4}{c}{\textbf{NIID}}                                                                                                                 \\ \cline{3-10} 
\multicolumn{1}{c|}{}                           &                                        & \multicolumn{1}{c|}{\textbf{PubMedQA}} & \multicolumn{1}{c|}{\textbf{FiQA}}  & \multicolumn{1}{c|}{\textbf{AQUA-RAT}} & \multicolumn{1}{c|}{\textbf{Mol-Instructions}} & \multicolumn{1}{c|}{\textbf{PubMedQA}} & \multicolumn{1}{c|}{\textbf{FiQA}}  & \multicolumn{1}{c|}{\textbf{AQUA-RAT}} & \textbf{Mol-Instructions} \\
\multicolumn{1}{c|}{}                           & MT-bench                               & \multicolumn{1}{c|}{Acc}               & \multicolumn{1}{c|}{Win\%}          & \multicolumn{1}{c|}{Acc}               & \multicolumn{1}{c|}{BertScore}                 & \multicolumn{1}{c|}{Acc}               & \multicolumn{1}{c|}{Win\%}          & \multicolumn{1}{c|}{Acc}               & BertScore                 \\ \hline
\textbf{FedAvg}         (oracle)                  & 4.475                                      & \multicolumn{1}{c|}{0.750}             & \multicolumn{1}{c|}{-}              & \multicolumn{1}{c|}{0.299}             & \multicolumn{1}{c|}{0.812}                     & \multicolumn{1}{c|}{0.747}             & \multicolumn{1}{c|}{-}              & \multicolumn{1}{c|}{0.252}             & 0.812                     \\
\textbf{FedAvg}                                 & -                                  & \multicolumn{1}{c|}{0.681}             & \multicolumn{1}{c|}{0.266}          & \multicolumn{1}{c|}{0.205}             & \multicolumn{1}{c|}{0.809}                     & \multicolumn{1}{c|}{0.664}             & \multicolumn{1}{c|}{0.354}          & \multicolumn{1}{c|}{0.205}             & 0.809                     \\ \hline
\textbf{FedAvg+PPL}                             & 4.525                                  & \multicolumn{1}{c|}{0.703}             & \multicolumn{1}{c|}{0.437}          & \multicolumn{1}{c|}{0.224}             & \multicolumn{1}{c|}{0.809}                     & \multicolumn{1}{c|}{0.684}             & \multicolumn{1}{c|}{0.544}          & \multicolumn{1}{c|}{0.217}             & 0.804                     \\
\textbf{FedAvg+DataInf}                         & 4.443                                  & \multicolumn{1}{c|}{0.728}             & \multicolumn{1}{c|}{0.457}          & \multicolumn{1}{c|}{0.224}             & \multicolumn{1}{c|}{0.811}                     & \multicolumn{1}{c|}{0.675}             & \multicolumn{1}{c|}{0.464}          & \multicolumn{1}{c|}{0.232}             & 0.807                     \\
\textbf{FedAvg+IFD}                             & 4.600                                  & \multicolumn{1}{c|}{0.714}             & \multicolumn{1}{c|}{0.622}          & \multicolumn{1}{c|}{0.244}             & \multicolumn{1}{c|}{0.812}                     & \multicolumn{1}{c|}{0.699}             & \multicolumn{1}{c|}{0.664}          & \multicolumn{1}{c|}{0.275}             & 0.815                     \\
\textbf{FedAvg+NUGGETS}                         & 4.443                                  & \multicolumn{1}{c|}{0.708}             & \multicolumn{1}{c|}{0.565}          & \multicolumn{1}{c|}{0.240}             & \multicolumn{1}{c|}{0.815}                     & \multicolumn{1}{c|}{0.682}             & \multicolumn{1}{c|}{0.566}          & \multicolumn{1}{c|}{0.232}             & 0.814                     \\ \hline
\textbf{FedDQC} & \textbf{4.780} & \multicolumn{1}{c|}{\textbf{0.751}} & \multicolumn{1}{c|}{\textbf{0.721}} & \multicolumn{1}{c|}{\textbf{0.290}} & \multicolumn{1}{c|}{\textbf{0.819}} & \multicolumn{1}{c|}{\textbf{0.751}} & \multicolumn{1}{c|}{\textbf{0.821}} & \multicolumn{1}{c|}{\textbf{0.280}} & \textbf{0.824} \\
\hline

\end{tabular}
}
\vspace{-0.6cm}
\label{tab:main}
\end{center}
\end{table*}

\noindent\textbf{Quality evaluation metric}
We propose the Instruction-Response Alignment (IRA) metric for data quality assessment, inspired by mutual information~\cite{kraskov2004estimating}, to evaluate the relevance between instructional prompts and responses using the global model. Specifically, IRA is computed locally with the global model to calculate the difference in inference loss between unconditioned responses and responses conditioned on their corresponding instructions. The following equation defines the scoring function $f_{IRA}$:
\begin{equation*}
    f_{IRA}((q^i,a^i)\in\mathcal{D}, \theta) = L(a^i;\theta) - L((a^i,q^i);\theta)
\end{equation*}

\vspace{-0.2cm}

\noindent where $L(a^i;\theta)=-\sum_{j=1}^{l_i} \log p(a^i_j|a^i_{<j};\theta)$ calculates the cross-entropy loss of generating response $a^i$ without given the instruction $q^i$, $L((a^i,q^i);\theta)$ is the cross-entropy loss given instruction $q^i$, which is defined in Section~\ref{Sec:IT_def}. $\mathcal{D}$ is dataset and $\theta$ represents model parameter for data quality evaluation. 

This metric subtly connects data quality with learning difficulty by reflecting how well the instruction aligns with the response, which in turn influences how easily the model can learn from the data. Visualization in Fig.~\ref{fig:datamap_c} supports this, see discussion in Section~\ref{sec:vis_part1}.

\noindent\textbf{Local dataset sort and select}
Using the efficient IRA, clients process the local dataset in two steps. First, clients sort their untrained local data in a descending order based on IRA values. Then client filter out low-quality samples using a global threshold $\lambda$, retaining only the high-quality data. 

Rather than solely relying on the pre-trained model to evaluate data quality, our approach incorporates re-scoring before each hierarchical federated training stage, dynamically adjusting the data selection process according to the model’s evolving capabilities throughout training. This adaptive approach offers a key advantage: better handling of low-quality data. As the model's abilities improve during training, it becomes more adept at distinguishing between challenging, valuable data and noisy or irrelevant data, making it more capable in selecting the right data and ensuring a more reliable and stable model. Section~\ref{sec:vis_part1} provides visualizations that offer strong evidence of the effectiveness of this methodology.

\vspace{-0.2cm}
\subsection{Training Stage: Hierarchical Training}
\vspace{-0.1cm}
\label{sec:train}

After controlling data quality in local dataset, the next step is federated training. In this stage, we propose quality-aware hierarchical training based on previous IRA scoring,  where models learn progressively from easier to harder data. This typically involves two steps:

\noindent\textbf{Step 1:} \textbf{Split to subsets:} For hierarchy $k$ in total $K$ hierarchies, the retained data for client $n$ is partitioned into $K-k$ separate hierarchies in a descending $\mathcal{H}_{nk}, \dots, \mathcal{H}_{nK}$ order with IRA values. Each hierarchy contains an equal number of samples, and any samples that have already been trained in previous hierarchies are removed. The subset $\mathcal{H}_{nk}$ with the highest score is selected as the training set for this hierarchical federated training. 

\noindent\textbf{Step 2:} \textbf{Federated training inside hierarchy:} During training, each clients choose the highest scored subset $\mathcal{H}_{nk}$ for training. By prioritizing high-quality, easy-to-learn data, the FL process starts with basic, highly relevant instruction-response samples, then gradually applies its instruction-following skills to more generalized tasks, and eventually progresses to solving more complex problems. This approach offers two key benefits: 1) it enables the model to build a strong foundational understanding, improving learning effectiveness and robustness; 2) it ensures consistent data quality in each training round, reducing the risk of divergence in the aggregated model. See visualization in Section~\ref{sec:vis_part2}

In each round of global aggregation and local updating, any federated algorithm could easily adapt to FedDQC. Section~\ref{sec:abl_flbaseline} demonstrate the effectiveness of FedDQC plugged in various FL algorithms.

\renewcommand{\algorithmiccomment}[1]{\hfill $\triangleright$ #1}


\vspace{-0.1cm}
\section{Disscussion}
\vspace{-0.1cm}
\label{sec:discussion}
\paragraph{Communication, privacy and computation}
"Privacy and communication efficiency are two primary concerns in FL".~\cite{advances} Our proposed FedDQC does not compromise on either of these aspects, as it does not introduce extra communication costs or privacy leakage, through training. Regarding computation, FedDQC adds only one step compared to FedAvg: scoring all the training data, which only requires inferencing rather than training. When keeping the batch size the same for training and inference, the scoring time accounts for approximately 1\% of the total training time. See Section \ref{abl:computation}.

\vspace{-0.1cm}
\section{Experiments}
\label{sec:experiments}
\vspace{-0.1cm}
\subsection{Experiment Setup}

\paragraph{Dataset and evaluation metric} We explore a real-world dataset and four task-specific datasets, PubMedQA~\cite{jin2019pubmedqa}, FiQA~\cite{yang2023fingpt}, AQUA-RAT~\cite{aqua} and Mol-Instructions~\cite{fang2023mol} covering diverse domains (i.e., medical, finance, math, and molecular science). To simulate real-world mixed-quality data, we introduced synthetic low-quality data at a 50\% proportion across the four domain-specific datasets. For more details please refer to Appendix~\ref{app:dataset}. 

\vspace{-0.2cm}
\paragraph{Models and training settings}
Our experiment is implemented on the OpenFedLLM~\cite{openfedllm} framework. We use LLama2-7b\cite{touvron2023llama} as the pre-trained model and adapt Low-Rank Adaptation (LoRA)~\cite{hu2021lora} to achieve fine-tuning. See Appendix ~\ref{app:train_setting}.

\vspace{-0.2cm}
\paragraph{Baselines}
We include four types of data quality evaluation metrics as data quality control baselines: perplexity (PPL)~\cite{de2022bertin}, loss, IFD~\cite{IFD}, NUGGETS~\cite{nuggets}, and DataInf~\cite{kwon2023datainf}. These four metrics are applied at the data-scoring stage. We select the high-score data for later federated training. 
In our experiments, DataInf and IFD are slightly adapted to federated scenarios, refer to Appendix~\ref{app:baseline} for more details.

\begin{table*}[t]
\centering
\vspace{-15pt}
\caption{Performance comparisons of random batching and two hierarchical training sequences with 5 quality evaluation metrics on PubMedQA in IID setting. IRA is a training-aware quality evaluation metric compatible with descending hierarchical training. The red box highlights the best result among all baselines, while the blue box highlights the best performance within the baseline.}
\vspace{-10pt}
\scalebox{0.69}{
\begin{tabular}{l|ccc|ccc|ccc|ccc}
\hline
\textbf{}           & \multicolumn{3}{c|}{\textbf{PubMedQA}}                                                             & \multicolumn{3}{c|}{\textbf{AQUA-RAT}}                                                               & \multicolumn{3}{c|}{\textbf{Mol-Instructions}}                                                                               & \multicolumn{3}{c}{\textbf{FiQA}}                                                                                                 \\
\hline
\textbf{Train order}& \textbf{random}                        & \textbf{ascend} & \textbf{descend}                       & \textbf{random}                         & \textbf{ascend} & \textbf{descend}                        & \textbf{random}                        & \textbf{ascend}                       & \textbf{descend}                       & \textbf{random}                        & \textbf{ascend}                       & \textbf{descend}                       \\
                \hline
\textbf{random}        & \multicolumn{3}{c|}{0.681}                                                                         & \multicolumn{3}{c|}{0.205}                                                                          & \multicolumn{3}{c|}{0.809}                                                                                               & \multicolumn{3}{c}{26.60}                                                                                                \\ \hline
\textbf{PPL} & \cellcolor[HTML]{E1EAFF}\textbf{0.703} & 0.663            & 0.685                                  & \cellcolor[HTML]{E1EAFF}\textbf{0.240} & 0.217           & 0.220                                  & \cellcolor[HTML]{E1EAFF}\textbf{0.809} & 0.809                                  & 0.807                                  & \cellcolor[HTML]{E1EAFF}\textbf{0.437} & 0.338                                  & 0.333                                  \\

\textbf{NUGGETS}    & \cellcolor[HTML]{E1EAFF}\textbf{0.708} & 0.682            & 0.674                                  & \cellcolor[HTML]{E1EAFF}\textbf{0.240} & 0.193           & 0.201                                  & \cellcolor[HTML]{E1EAFF}\textbf{0.815} & 0.814                                  & 0.810                                   & 0.457                                  & \cellcolor[HTML]{E1EAFF}\textbf{0.681} & 0.320                                  \\
\textbf{IFD}        & \cellcolor[HTML]{E1EAFF}\textbf{0.714} & 0.697            & 0.656                                  & \cellcolor[HTML]{E1EAFF}\textbf{0.244} & 0.217           & 0.193                                  & 0.814                                  & \cellcolor[HTML]{E1EAFF}\textbf{0.820} & 0.799                                  & \cellcolor[HTML]{E1EAFF}\textbf{0.622} & 0.612                                  & 0.287                                  \\
\textbf{DataInf}    & \cellcolor[HTML]{E1EAFF}\textbf{0.728} & 0.720            & 0.717                                  & \cellcolor[HTML]{E1EAFF}\textbf{0.224} & 0.181           & 0.169                                  & \cellcolor[HTML]{E1EAFF}\textbf{0.811} & 0.806                                  & 0.810                                   & \cellcolor[HTML]{E1EAFF}\textbf{0.565} & 0.223                                  & 0.300                                  \\
\textbf{IRA}        & 0.725                                  & 0.718            & \cellcolor[HTML]{FFCCC9}\textbf{0.751} & 0.252                                  & 0.197           & \cellcolor[HTML]{FFCCC9}\textbf{0.290} & 0.817                                  & 0.803                                  & \cellcolor[HTML]{FFCCC9}\textbf{0.819} & 0.690                                 & 0.432                                  & \cellcolor[HTML]{FFCCC9}\textbf{0.721} \\ \hline
\end{tabular}}
\label{tab:train_seq}
\vspace{-10pt}

\end{table*}

\vspace{-0.1cm}
\subsection{Main result}
We conduct experiments on a real-world dataset and four synthetic domain-specific datasets with synthetic low-quality data on both IID and NIID settings, shown in Table \ref{tab:main}.

\vspace{-0.2cm}
\paragraph{Applicability on synthetic dataset}
We compare FedAvg with the original dataset (referred as oracle in table), and the synthetic mixed-quality dataset,  applying 4 data selection baselines and the FedDQC. For FiQA datasets, all win rate are compared with high-quality data with FedAvg in both settings. To ensure fairness, we adjust the global threshold $\lambda$ to keep the number of training samples consistent and maintain the same number of training rounds. Key observations from Table~\ref{tab:main} include: 1) FedDQC consistently mitigates the impact of low-quality data and outperforms other baseline. 2) In some cases, FedDQC even surpasses the performance of training on fully clean data, benefiting from progressive training and the fact that not all data in the oracle dataset is equally valuable.

\vspace{-0.2cm}
\paragraph{Applicability on real-world dataset}
We present the results on the real-world dataset, Fed-WildChat, in Table~\ref{tab:main}, where various data scoring metrics were applied to select 70\% of the oracle training data for the same number of training rounds. We can see that: 1) FedDQC outperforms all other baselines and even surpasses the full dataset training performance, indicating the presence of low-quality data in the real-world dataset. 2) Data selection methods based on DataInf and NUGGEST perform worse than using the full dataset. This suggests that using gradients for data attribution in real-world datasets, as in the case of DataInf, is challenging. Additionally, the diverse distribution of real-world data makes it difficult to evaluate data quality using a fixed validation set, as shown by the performance of NUGGEST.

\noindent\textbf{Applicability on different FL algorithms}
\label{sec:abl_flbaseline}
We combine FedDQC with several FL algorithms beyond FedAvg, including FedAvgM~\cite{fedavgm}, FedAgrad~\cite{fedyogi}, FedYOGI~\cite{fedyogi}, and FedAdam~\cite{fedyogi}. Table~\ref{tab:fedbaseline} shows that FedDQC significantly boosts performance across these algorithms. For example, in the mix-quality scenario, FedAdagrad's performance improved from 0.709 to 0.731 with FedDQC, illustrating the effectiveness of FedDQC in enhancing model performance when paired with other FL algorithms.

\begin{table}[t]
\caption{Compatability with other 4 federated algorithms on PubMedQA, IID setting. The last line shows the improvement on mixed-quality data with FedDQC added.}
\centering
\vspace{-10pt}
\scalebox{0.6}{
\begin{tabular}{l|c|c|c|c|c}
\toprule
            & \textbf{FedAvg}         & \textbf{FedAvgM}        & \textbf{FedAdagrad} & \textbf{FedYOGI} & \textbf{FedAdam} \\\midrule
\textbf{oracle}      & 0.750          & 0.732          & 0.717    & 0.512   & 0.527   \\
\textbf{mix-quality} & 0.681          & 0.676          & 0.709    & 0.498   & 0.476   \\
\textbf{+FedDQC}     & 0.751 & 0.729 & 0.731    & 0.512   & 0.531  \\ 
 & \textbf{(+7\%)} & \textbf{(+5.3\%)} & \textbf{(+2.2\%)} & \textbf{(+1.4\%)} & \textbf{(+5.5\%)} \\
\bottomrule
\end{tabular}}
\vspace{-0.5cm}
\label{tab:fedbaseline}
\end{table}

\vspace{-0.1cm}

\begin{figure*}[t]
    \vspace{-0.5cm}
    \subfigure[Ground truth]{   
        \begin{minipage}{4.9cm}
        \centering    
        \vspace{-5pt}
        \includegraphics[scale=0.2]{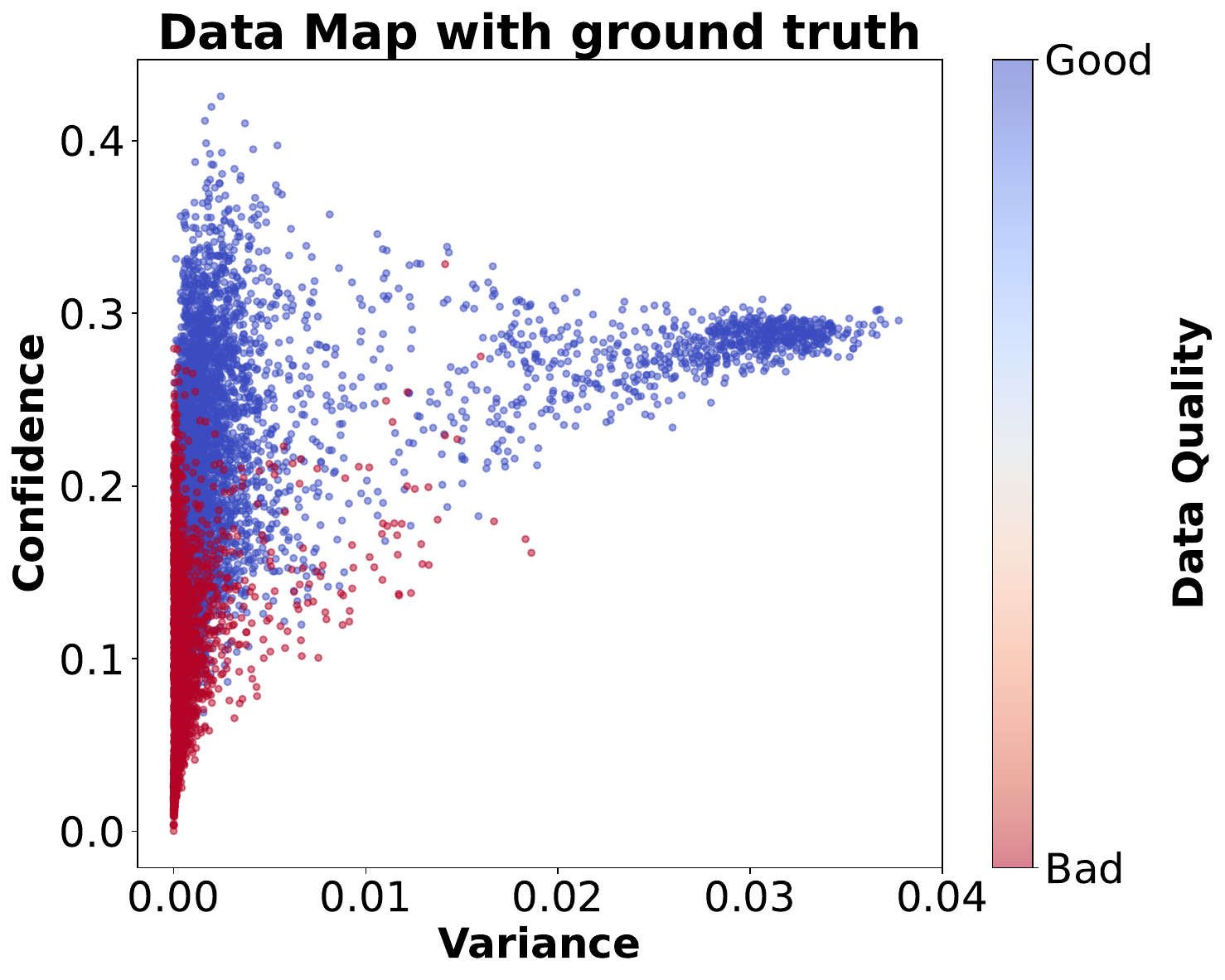}  
        \end{minipage}
        \label{fig:datamap_a}
    }
    \subfigure[IRA score on pre-trained model]{
    \vspace{-0.2cm}
        \begin{minipage}{5.1cm}
            \centering
            \vspace{-5pt}
            \includegraphics[scale=0.21]{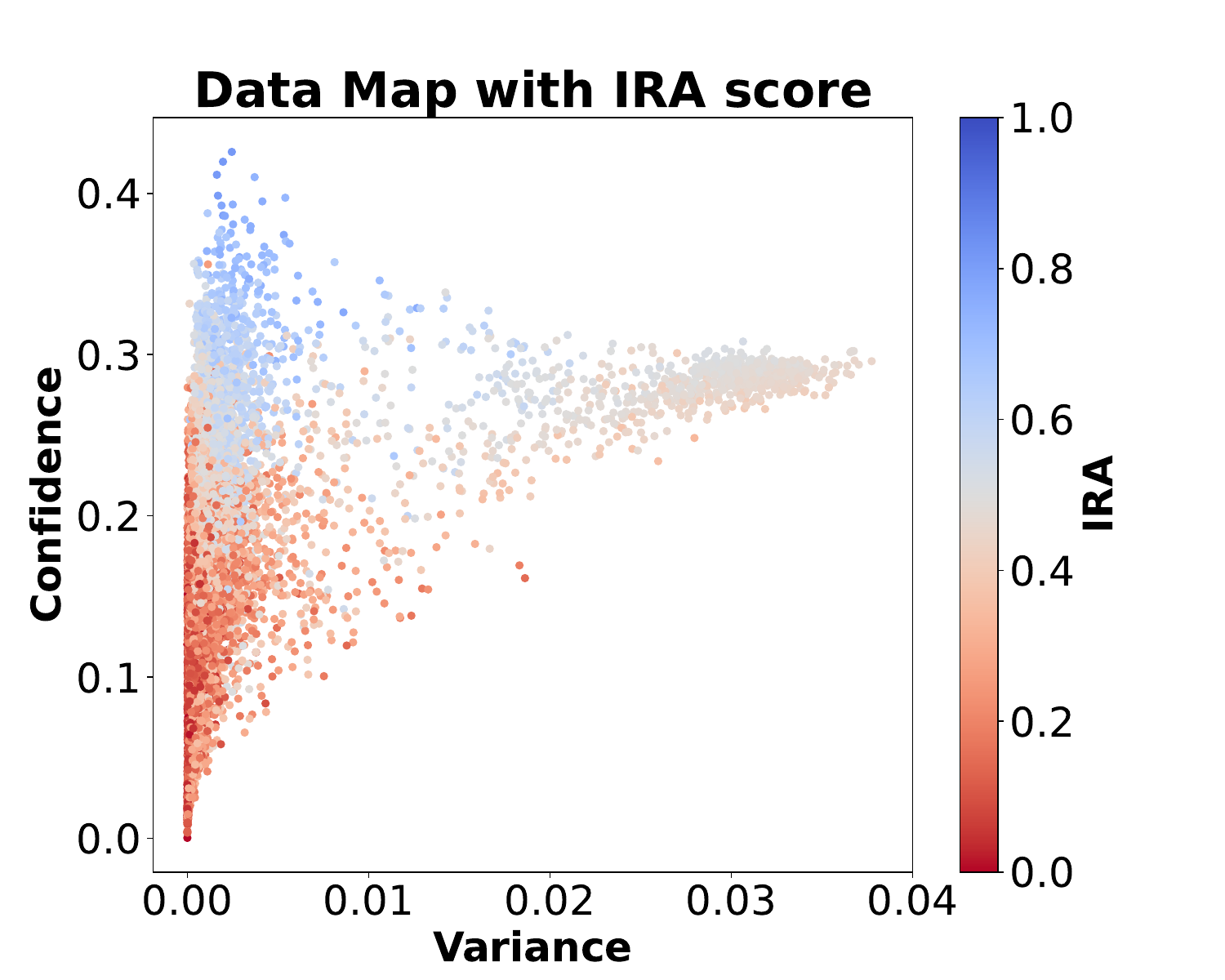}
        \end{minipage}
        \label{fig:datamap_b}
    }
    \subfigure[IRA score on finetuned model]{
        \vspace{-0.2cm}
        \begin{minipage}{4.8cm}
            \centering
            \vspace{-5pt}
            \includegraphics[scale=0.21]{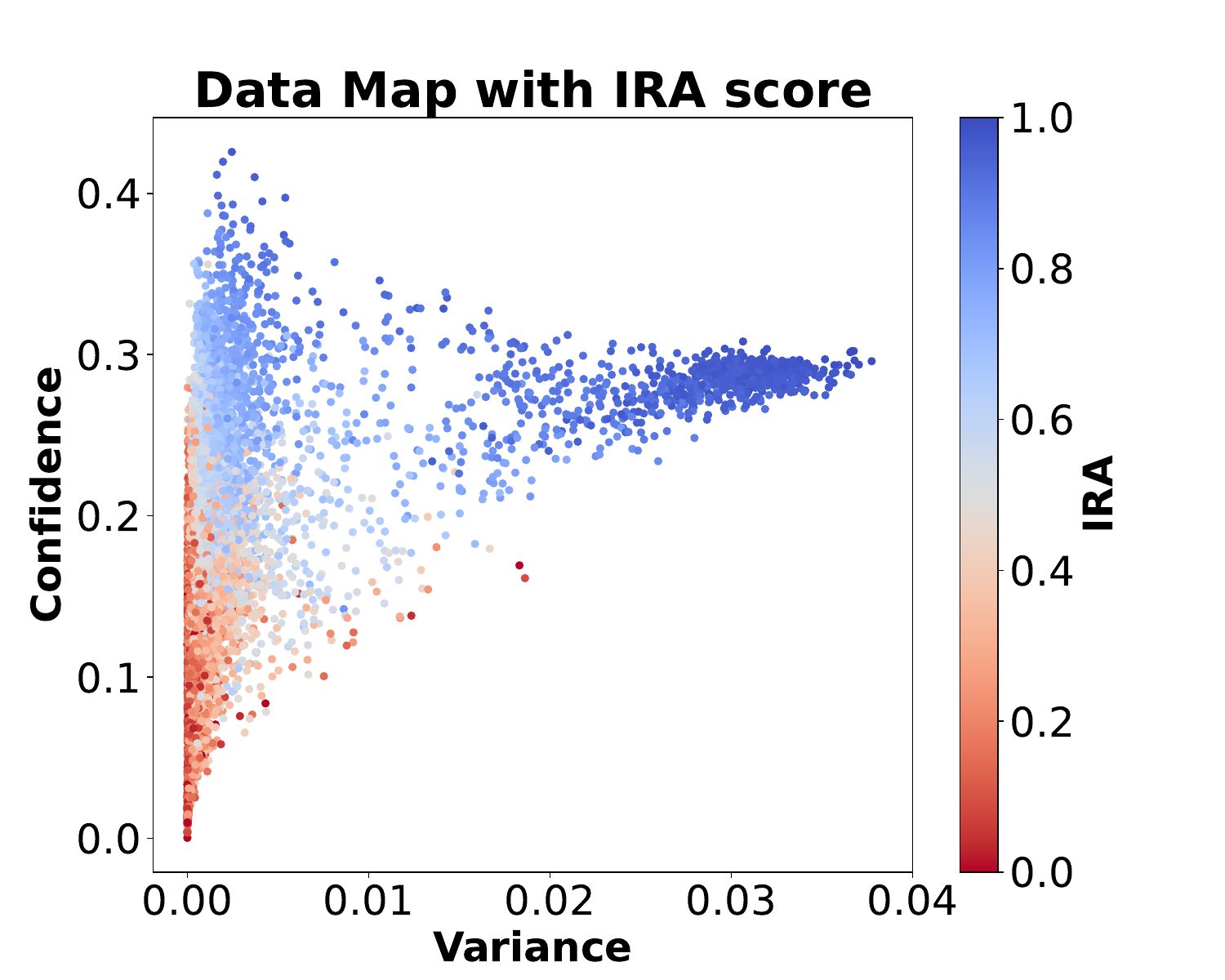}
        \end{minipage}
        \label{fig:datamap_c}
    }
    \vspace{-0.5cm}
    \caption{Data Map visualization. (a) Data Map with ground truth quality label. (b) Data Map with IRA scores on a pre-trained model. (c) Data Map with IRA scores on fine-tuned model.}
    \vspace{-0.5cm}
    \label{fig:datamap}
\end{figure*}

\vspace{-0.2cm}
\subsection{Visualization}
\vspace{-0.1cm}
\label{sec:vis}
Data Map~\cite{datamap} is a data training dynamics visualization tool, which tracks each sample's inference probability across training epochs. The Confidence (y-axis) is the mean of these probabilities and the Variance (x-axis) is the variance of these probabilities. Fig~\ref{fig:datamap} shows 5 types of low-quality data (noisy token, deleted token, truncation, and swapped responses) in a centralized setting on LLaMA-2-7b~\cite{touvron2023llama} model and dataset PMC-Llama~\cite{wu2024pmc} with 8000 samples, of which 50\% are low-quality samples.

\vspace{-0.1cm}
\paragraph{Relation of IRA and training difficulty} 
\label{sec:vis_part1}
IRA metric subtly connects data quality with learning difficulty. Fig.~\ref{fig:datamap_b} shows how this evaluation method closely reflects the relationship between IRA scores and the dynamic process of data during training. The Data Map in Fig.~\ref{fig:datamap_a} reveals a clear pattern between data quality and its training dynamics. For instance, data with high confidence and high variance are easier for the model to learn and perform better on, while low-variance, low-confidence data, like those in the bottom-left corner, are harder to learn and represent low-quality data. As Fig.~\ref{fig:datamap_b} illustrated, IRA aligns well with this dynamic tracking approach: 1) high-scoring data are easier to learn, with higher variance and lower confidence; 2) low-scoring data tend to cluster in the lower-left corner, with both lower confidence and variance, indicating more difficulty in learning, making them more likely to be noisy or irrelevant and negatively impacting model performance.

\vspace{-0.1cm}
\paragraph{Effectiveness of iterative scoring} 
\label{sec:vis_part2}
As shown in Fig.\ref{fig:datamap_c}, the model re-scored after training distinguishes data quality more clearly than the pre-trained model in Fig.\ref{fig:datamap_b}. Notably, high-scoring data tend to appear in regions with higher variance, indicating that the model is more confident in these challenging, yet informative samples.

\vspace{-0.1cm}
\subsection{Emperical analysis of FedDQC}

\subsubsection{The effectiveness of hierarchical training}
\vspace{-0.1cm}
To demonstrate the close integration of IRA scores with hierarchical training, we compared 3 training sequences: random, ascending, and descending; across 4 domain-specific datasets in the IID setting, as shown in Table~\ref{tab:train_seq}. The experiments reveal that: 1) IRA’s relationship with easy-to-hard hierarchical training is mutually reinforcing, with descending sequence training significantly improving IRA-based data selection across all datasets. Notably, IRA consistently outperforms other quality evaluation metrics, regardless of the training sequence. 2) Other quality metrics do not consistently benefit from hierarchical training, highlighting their incompatibility with this approach.

\begin{figure}
    \centering
    \vspace{-0.1cm}
    \includegraphics[scale=0.3]{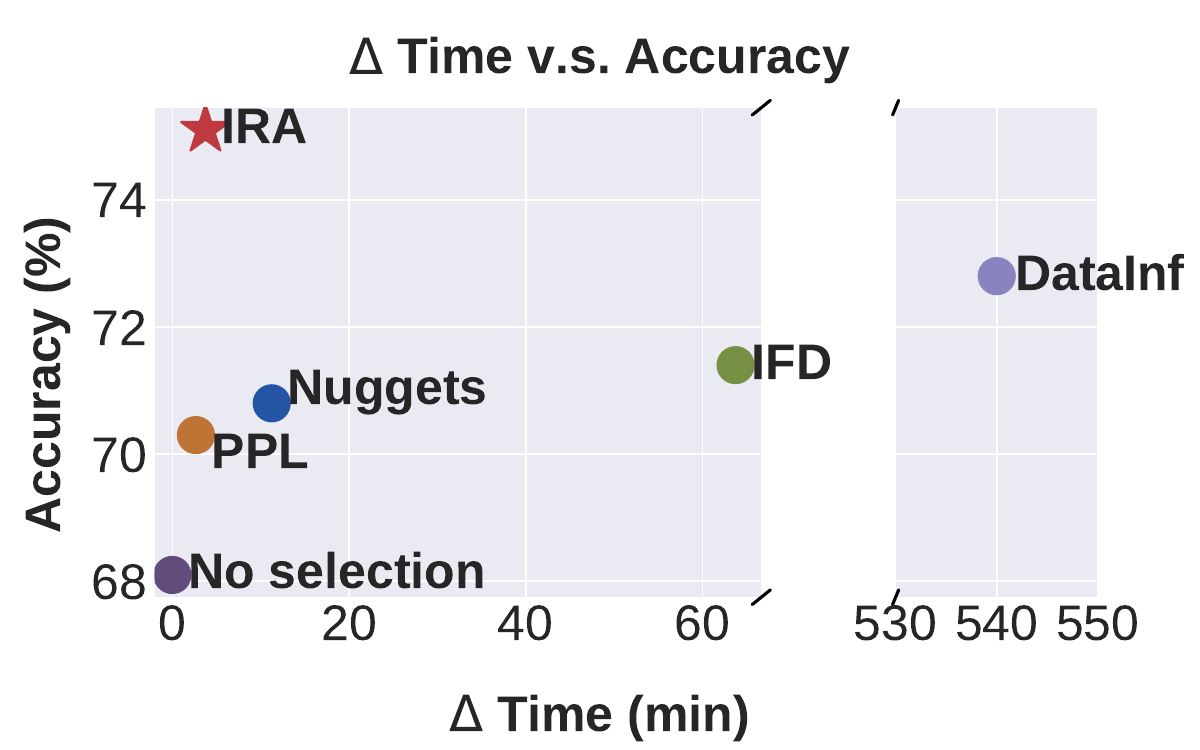}
    \vspace{-0.3cm}
    \caption{Comparison of additional computation costs and performance gain after applying to different quality evaluation metrics on PubMedQA IID setting. IRA adds minimal computational overhead while significantly improves performance by data quality control.}
    \label{fig:computation}
    \vspace{-0.5cm}
\end{figure}

\vspace{-0.2cm}
\subsubsection{Computational analysis}
\label{abl:computation}
We evaluated the additional computational costs of four data quality evaluation metrics compared to IRA during the data scoring stage, alongside their training performance on the PubMedQA dataset under an IID setting in Figure~\ref{fig:computation}. The experiment shows that: 1) compared to the total training time in FedAvg, 300.6 minutes, IRA only takes 1\% training time for data scoring, making it scalable for large datasets; 2) Compared to PPL, which is too simple to be effective. IRA uses an extra 1 minute, around 0.3\% training time, for scoring than PPL but has much higher performance;
3) Compared to the second well-performed metric, DataInf, IRA takes extremely less time, around 1/150 of the scoring time than DataInf. In conclusion, IRA is a computationally efficient, scalable data quality measuring metric, greatly enhancing data quality control.

\vspace{-0.2cm}
\subsubsection{Data quality impact analysis}
\begin{figure}[t]
\centering
\hspace{-20pt}
\subfigure[PubMedQA]{   
\begin{minipage}{3.6cm}
\centering    
\includegraphics[scale=0.21]{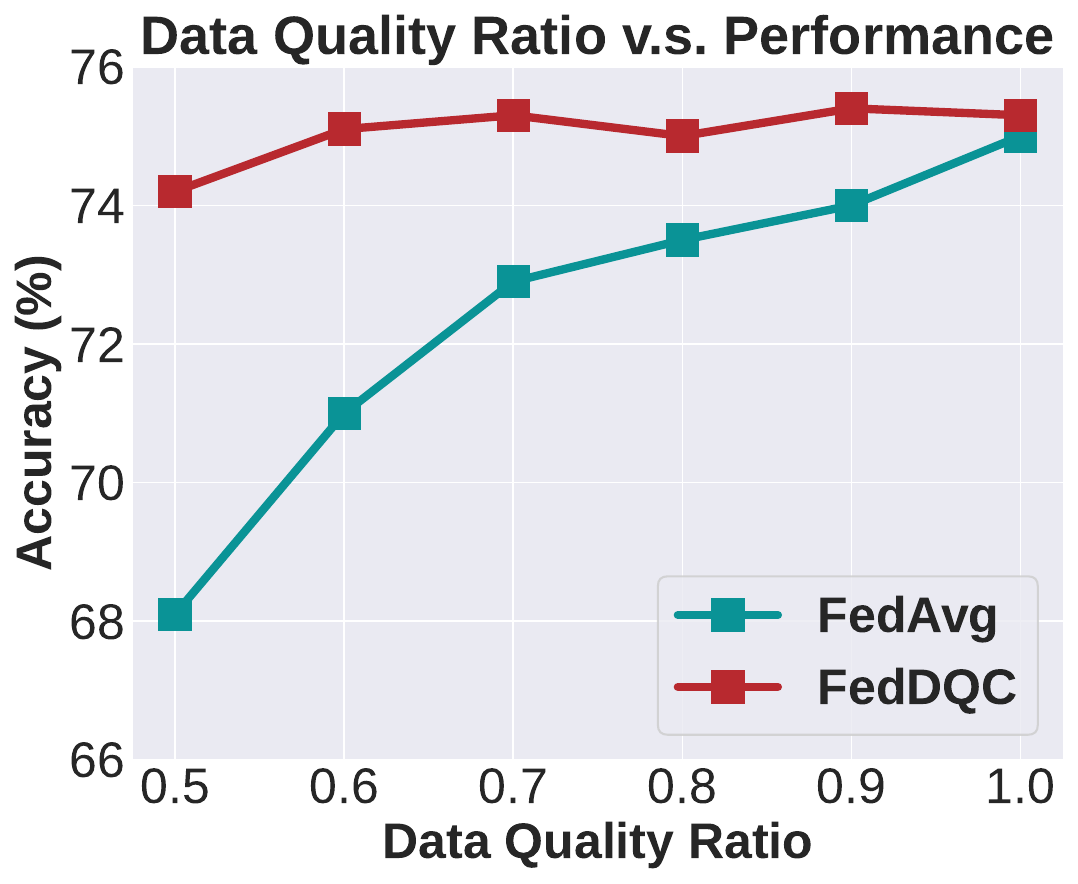}  
\end{minipage}
}
\subfigure[FiQA]{ 
\begin{minipage}{3.6cm}
\centering
\includegraphics[scale=0.21]{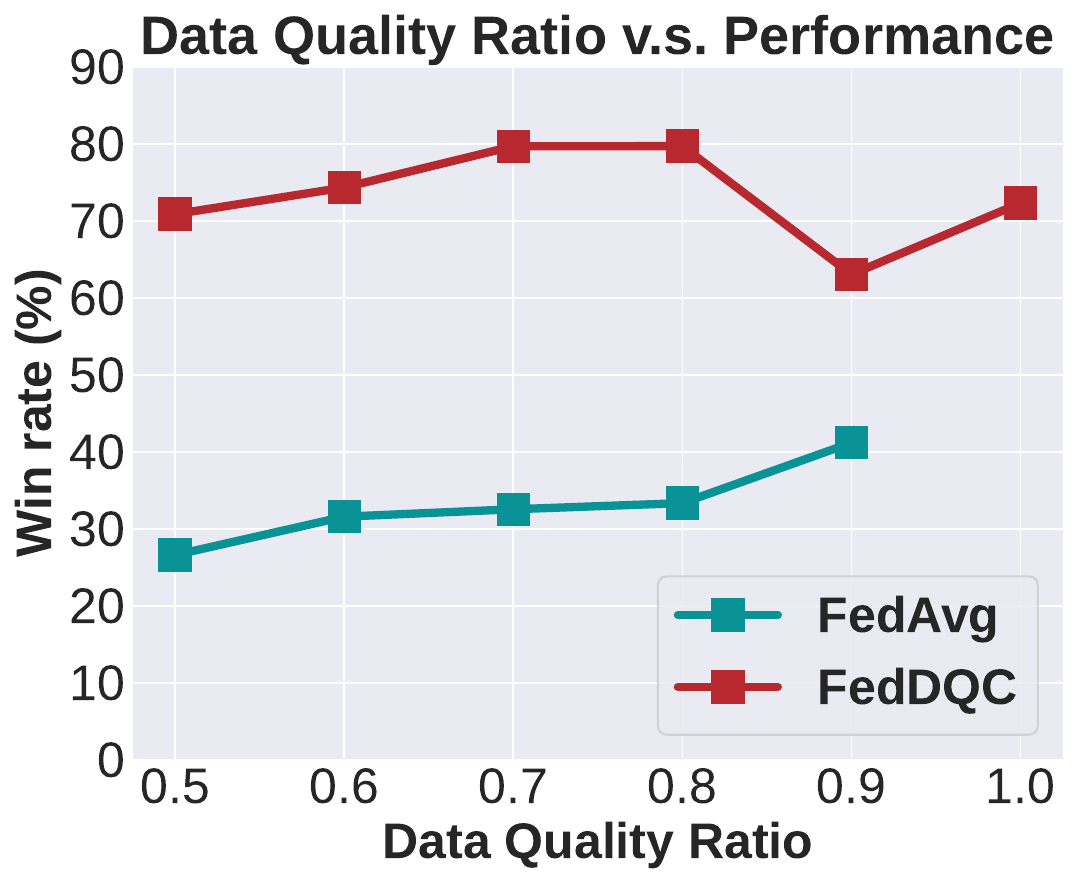}
\end{minipage}
}

\hspace{-20pt}
\subfigure[AQUA\_RAT]{ 
\begin{minipage}{3.6cm}
\centering
\vspace{-5pt}
\includegraphics[scale=0.2]{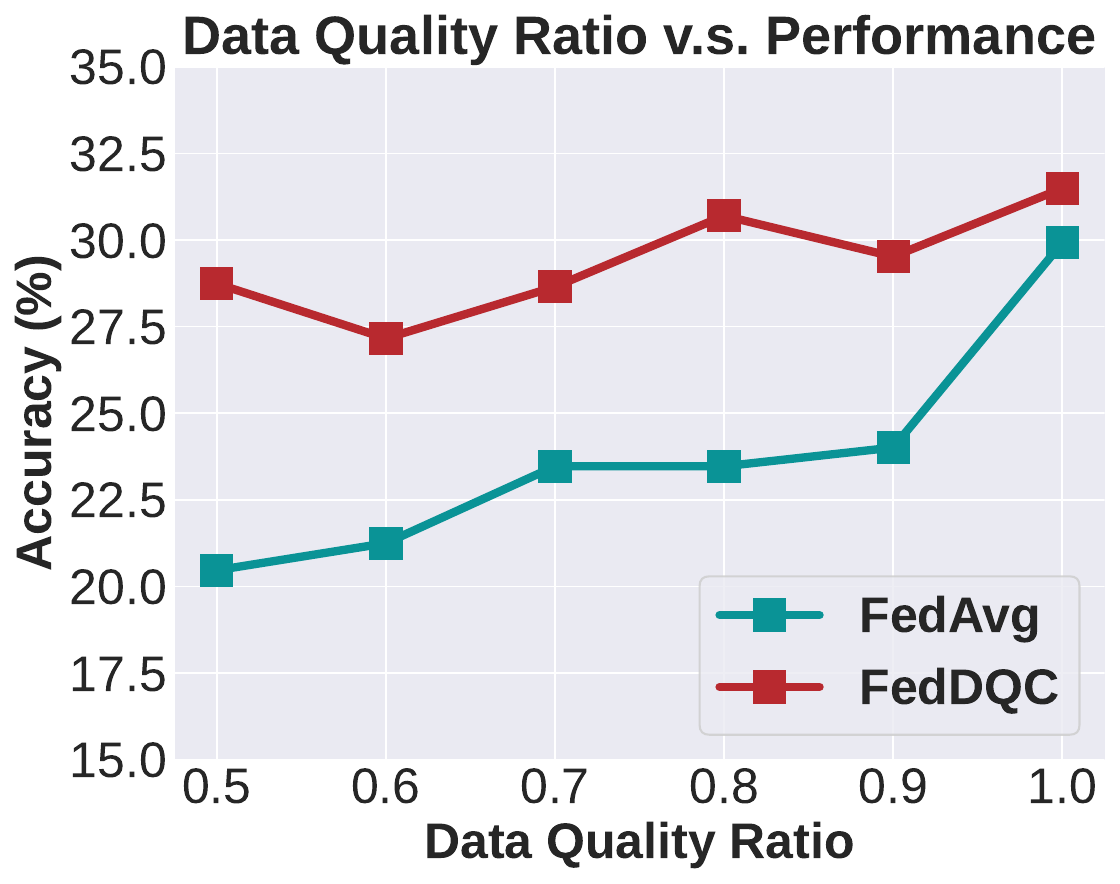}
\end{minipage}
}
\subfigure[Mol-Instructions]{ 
\begin{minipage}{3.6cm}
\centering
\vspace{-5pt}
\includegraphics[scale=0.2]{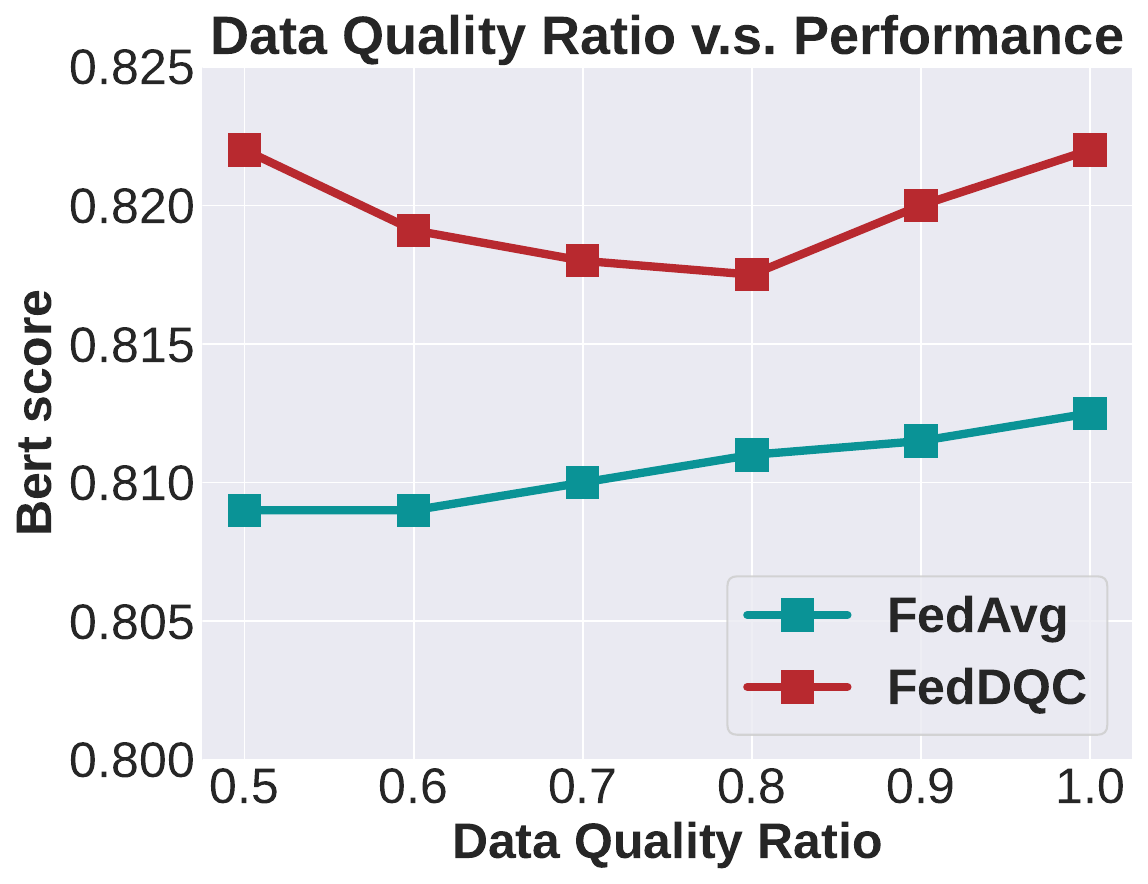}
\end{minipage}
}
\vspace{-10pt}
\caption{Comparison of FedAvg and FedDQC in various data quality ratios. (a)-(d) show the performance under different data quality ratio on various datasets. FedDQC is consistently higher than FedAvg in all data quality ratio on four datasets. 
}
\vspace{-0.5cm}
\label{fig:quality_ratio_performance}
\end{figure}

To examine how data quality impacts training, we quantify the dataset's overall quality as the ratio of aligned data to total data, and conduct experiments with varying data quality ratios (0.5 to 1.0) across four domain-specific datasets in the IID setting. For FiQA, we use win rates compared to the original dataset trained with FedAvg, so we exclude the 1.0 quality ratio for FedAvg. Key observations from Fig~\ref{fig:quality_ratio_performance} include: 
1) FedAvg performance decreases as the data quality ratio drops, showing the significant impact of low-quality data on training. 
2) FedDQC outperforms FedAvg in all quality ratio settings, demonstrating the robustness of its data quality control. 
3) Even with a quality ratio of 1.0 (no synthetic low-quality data), FedDQC consistently outperforms FedAvg, indicating its effectiveness in enhancing training performance, even in non-synthetic datasets.

\begin{figure}
    \centering
    \subfigure[FedAvg]{
    \begin{minipage}{3.6cm}
    \centering
        \includegraphics[scale=0.15]{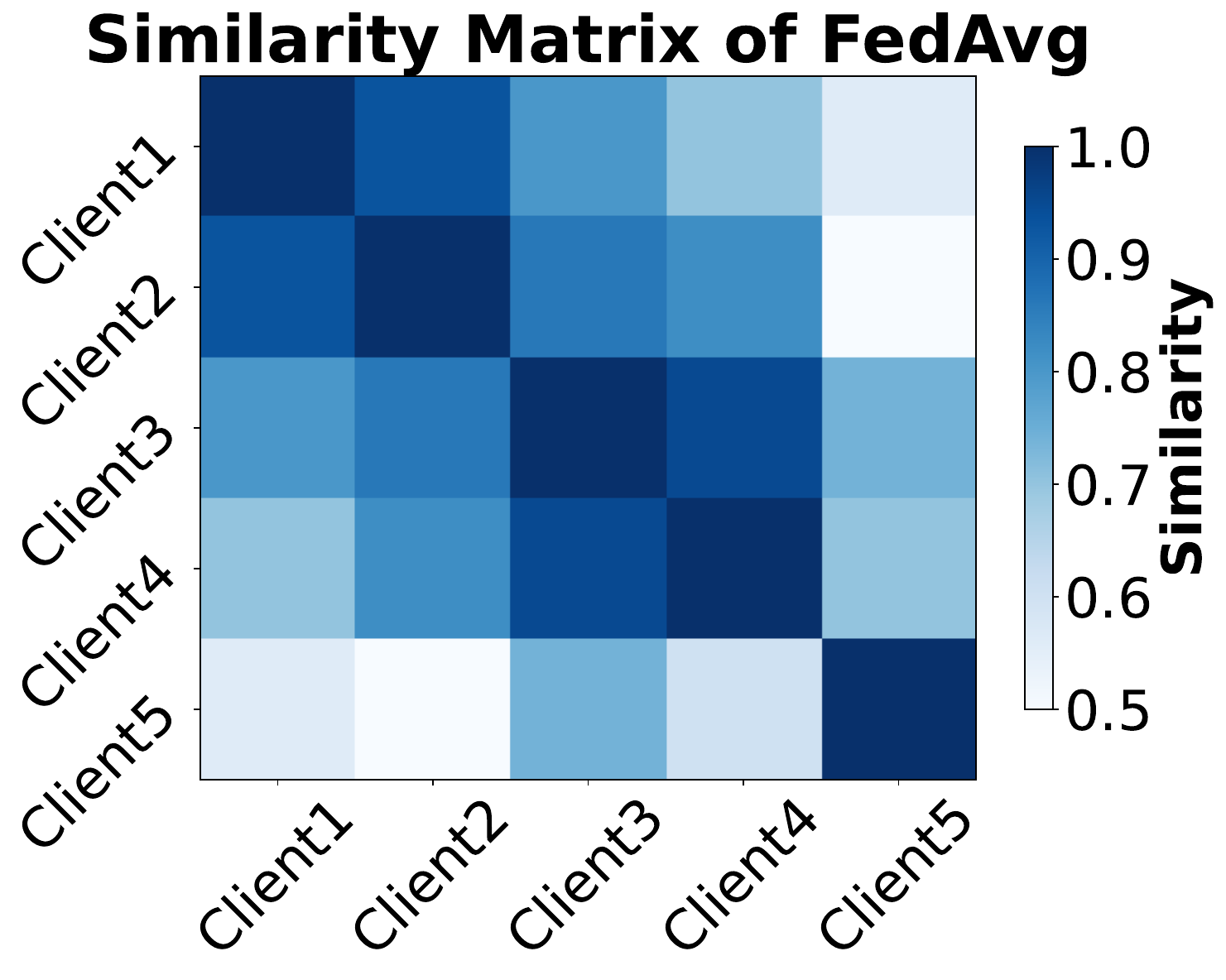}
    \end{minipage}
    }
    \subfigure[FedDQC]{
    \begin{minipage}{3.6cm}
    \centering
        \includegraphics[scale=0.15]{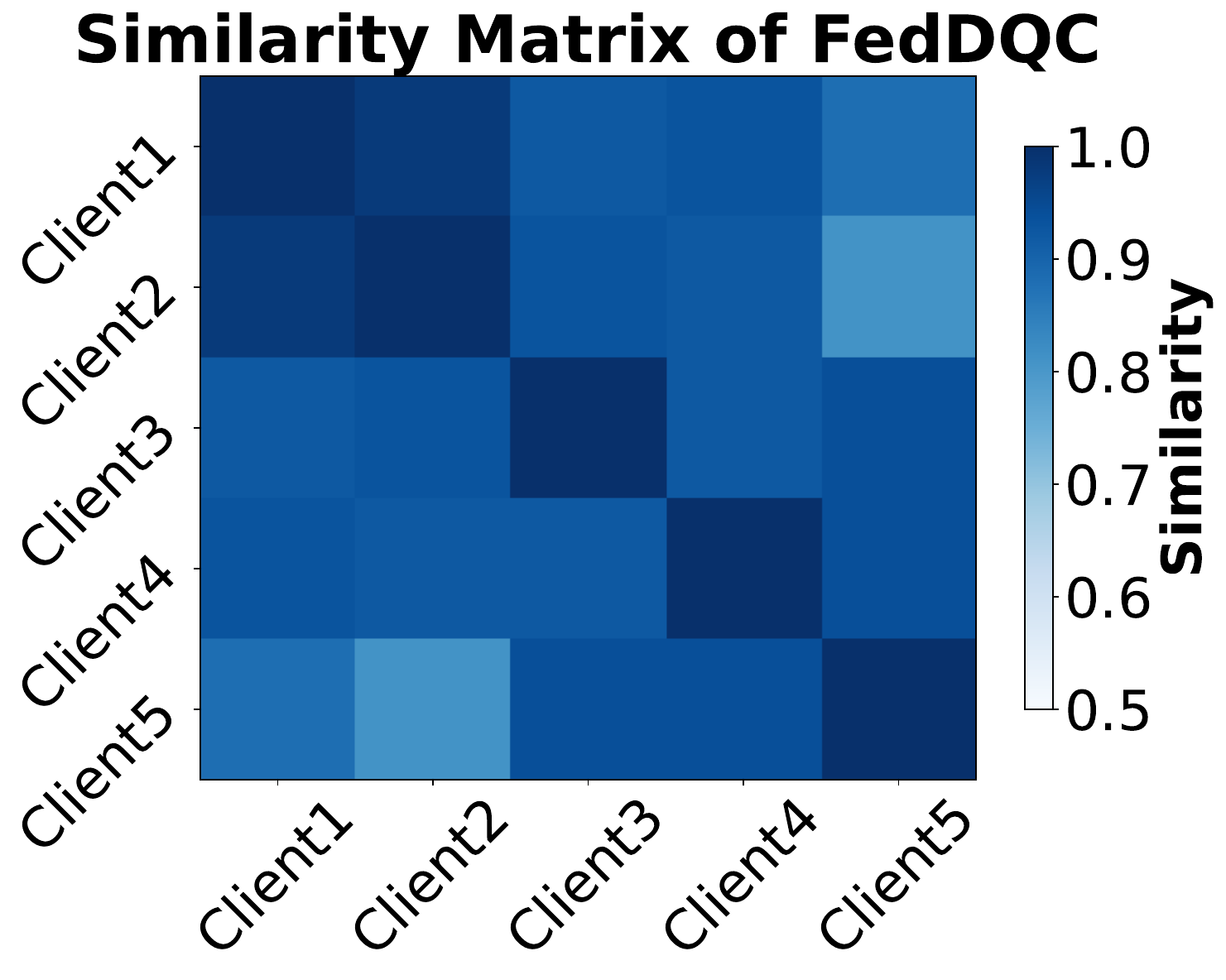}
    \end{minipage}
    }
    \vspace{-15pt}
    \caption{Model similarity comparison between FedAvg and FedDQC on PubMedQA dataset in NIID settings.}
    \label{fig:sim}
    \vspace{-0.5cm}
\end{figure}

\begin{figure*}[t]
\centering
\hspace{-4mm}
\subfigure[Global threshold $\lambda$]{   
\begin{minipage}{4.8cm}
\centering
\includegraphics[scale=0.22]{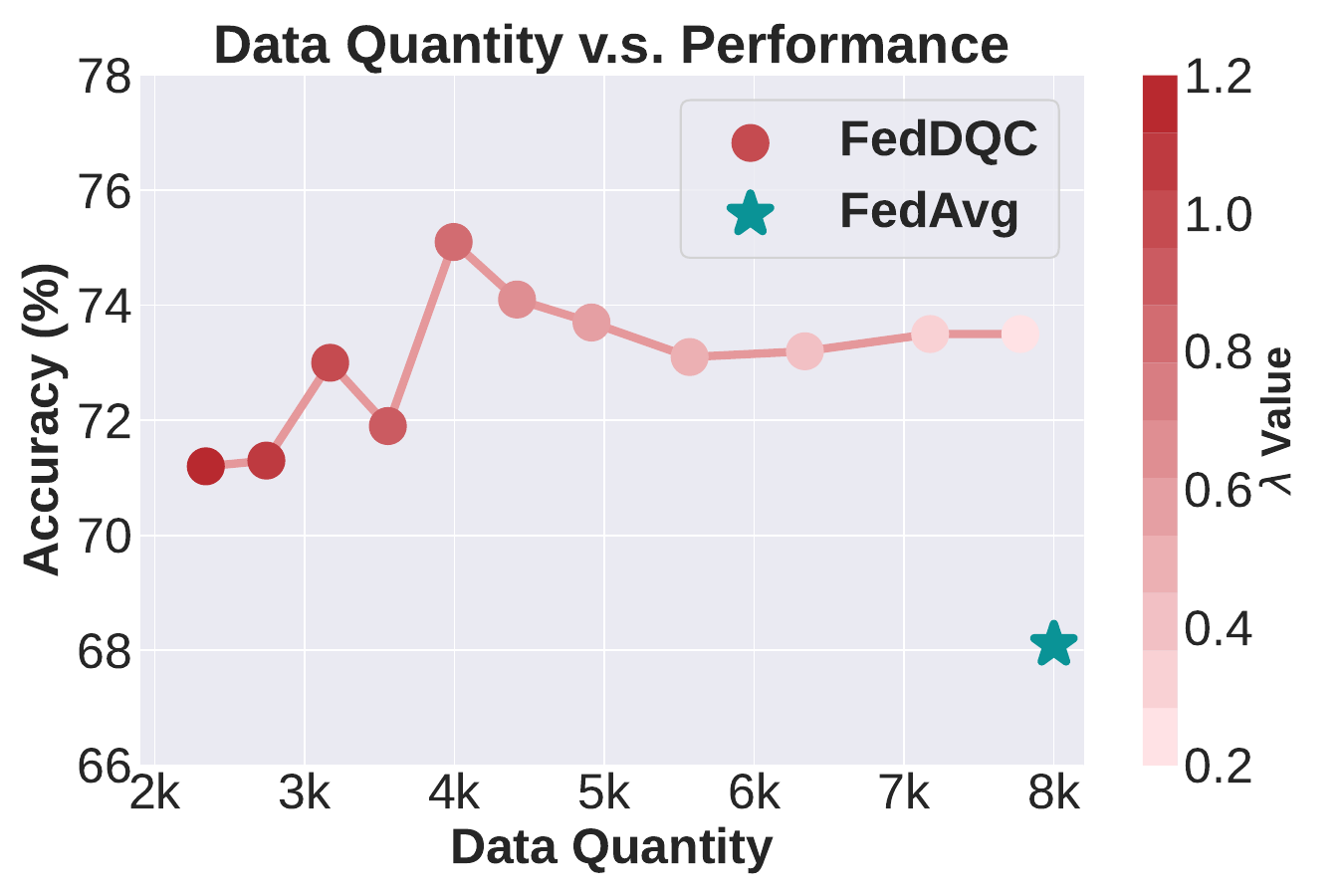}  
\end{minipage}
\label{subfig:threshold}
}
\label{fig:thres}
\subfigure[Global thre. v.s. Data quality ratio]{ 
\begin{minipage}{4.8cm}
\centering

\includegraphics[scale=0.23]{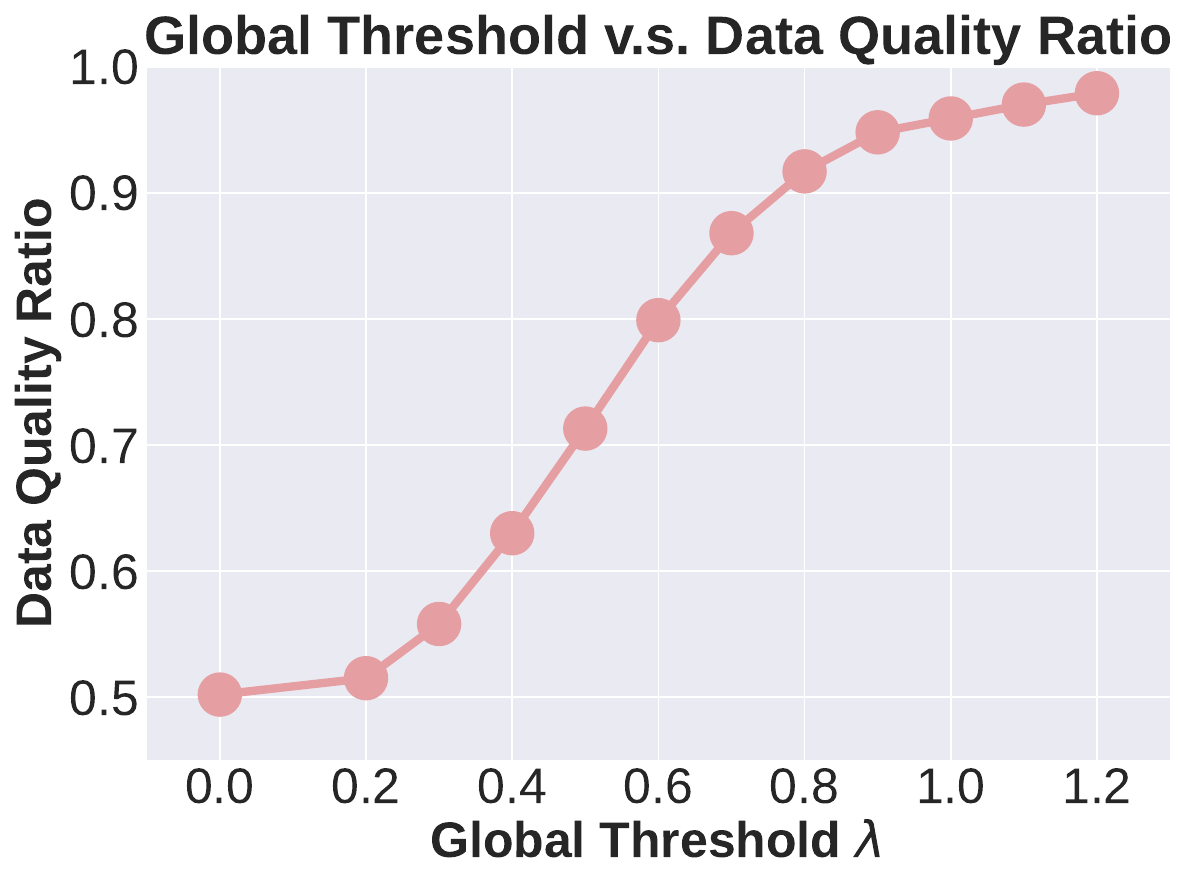}
\end{minipage}
\label{subfig:data_quality_ratio}
}
\subfigure[Number of hierarchies $K$]{ 
\begin{minipage}{4.8cm}
\centering

\includegraphics[scale=0.23]{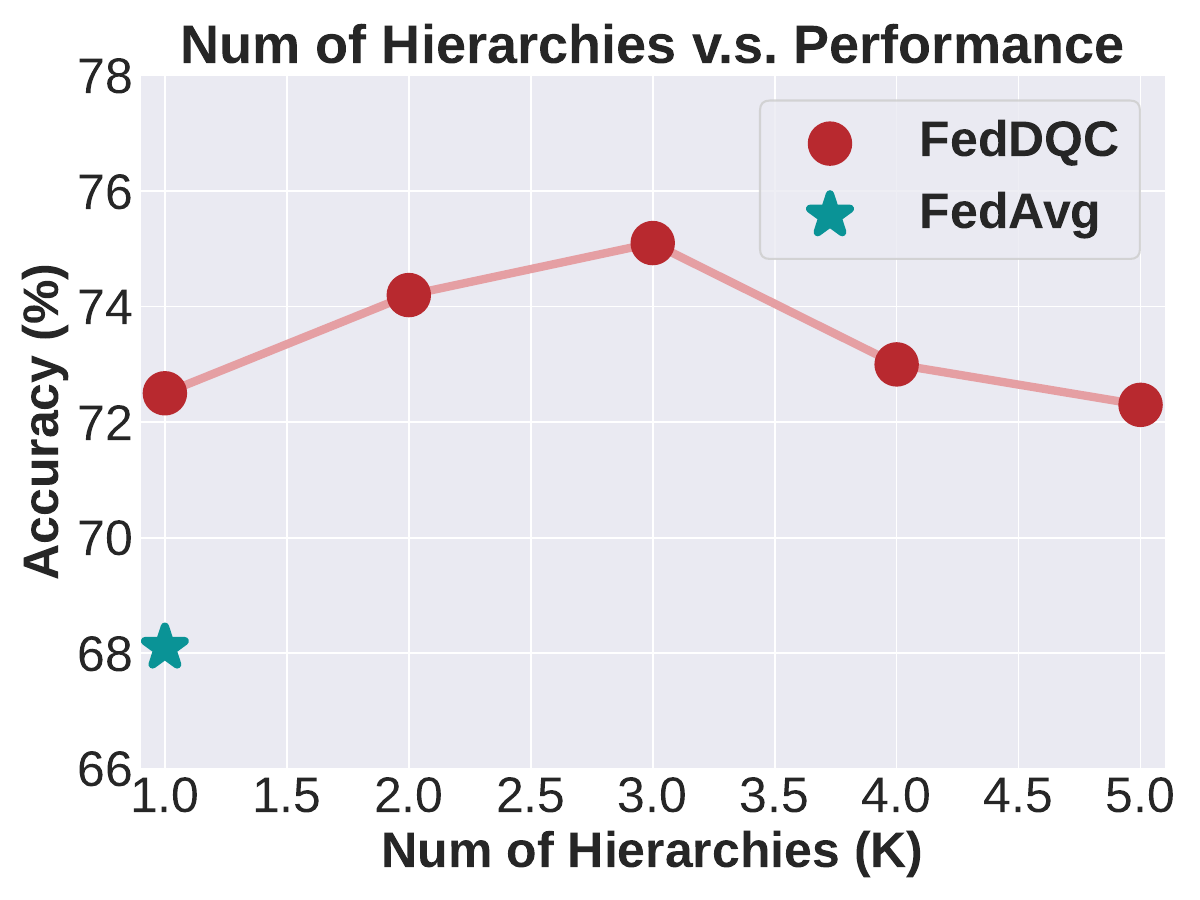}
\end{minipage}
\label{subfig:num_hierarchy}
}
\vspace{-0.5cm}
\caption{Ablation study. (a) Effect of global threshold on overall data quantity and training performance of FedDQC. Experiments show that FedDQC is robust to the global threshold. (b) Effects of global threshold on the quality ratio of all training data. (c) The effect of various hierarchies on training performance in FedDQC training.}
\end{figure*}

\vspace{-0.1cm}
\subsubsection{Convergence and Model Similarity analysis}
To demonstrate FedDQC's impact on convergence and data heterogeneity, we compared local model similarity at round 50 between FedAvg and FedDQC in a quality NIID setting with 5 clients. Fig.~\ref{fig:sim} shows that, in FedAvg, model similarity is generally low due to data quality differences, particularly for client 5 with lower-quality data. In contrast, FedDQC’s hierarchical training improves model similarity by filtering out low-quality data, reducing its negative impact, and enhancing aggregation. This results in a more stable global model, minimizing data heterogeneity and improving performance in heterogeneous settings.

\vspace{-0.1cm}
\subsection{Hyperparameter ablation}

\paragraph{Global threshold}
To demonstrate the threshold robustness of FedDQC, we further examine the impact of the global threshold $\lambda$ on the PubMedQA dataset with the IID setting.  As shown in Fig.~\ref{subfig:threshold}, the performance of FedDQC remains stable across varying $\lambda$ values, indicating its insensitivity to the threshold. Even with changing data quantities, FedDQC consistently outperforms FedAvg, demonstrating its robustness. Additionally, as the threshold decreases, the data quality ratio in the selected data increases, see Figure~\ref{subfig:threshold}~\ref{subfig:num_hierarchy}, but performance is more sensitive to the total data quantity than to data quality. This is evident from the asymmetric performance drop around 4k training data, where a decrease in data quantity results in a more pronounced performance decline.

\vspace{-0.1cm}
\paragraph{Number of hierarchies} 
Under the IID setting on PubMedQA, we tune the number of hierarchies in FedDQC $K\in\{1, 2, 3, 4, 5\}$. Figure~\ref{subfig:num_hierarchy} show that: 1) $K=3$ is optimal; 2) Beyond $K=3$ further increasing the number of hierarchies leads to a slight decline in accuracy. This suggests that while hierarchical training enhances learning, too many hierarchies may reduce diversity, slightly hindering performance.

\vspace{-0.1cm}
\section{Conclusions}
\vspace{-0.1cm}
\label{sec:conclusion}
In this paper, we introduce FedDQC, a novel framework for data quality control in federated instruction-tuning of LLMs. FedDQC combines a new data quality assessment metric (IRA) with federated hierarchical training, where data quality is dynamically evaluated during training. Our experiments demonstrate that FedDQC adds minimal computational overhead while significantly improving model performance. The integration of IRA, adaptive scoring, and hierarchical training enhances both efficiency and robustness, making FedDQC a promising approach for effective controlled data quality in mixed-quality distributed scenario.

\clearpage
\section{Limitations}
A limitation of this study is that it assumes all local models share the same architecture, which is achievable when fine-tuning with the LoRA adapter. However, this approach may not be suitable for scenarios involving different local model architectures. Additionally, the study does not address the integration of data diversity into the design.

\bibliography{custom}

\appendix

\clearpage

\section{Appendix}

\begin{table*}[t]
\centering
\caption{Dataset information and evaluation metrics}
\label{tab:dataset}
\begin{tabular}{lccccc}
\toprule
Dataset  & Evaluation metrics & Domain & $\#{samples}$ & $\hat{L}_{inst.}$ & $\hat{L}_{Resp.}$\\
\midrule
PubMedQA~\cite{jin2019pubmedqa}     & Acc        & medical   & 211 k & 471.1 & 71.4 \\
FiQA~\cite{yang2023fingpt}         & Win rate  & financial & 17.1 k & 42.1 & 255.7 \\
AQUA-RAT~\cite{aqua}     & Acc        & math      & 97.5 k & 77.4 & 105.7 \\
Mol-Instructions~\cite{fang2023mol} & Bert score & molecular & 38 k & 110.5 & 107.8 \\
Alpaca-GPT4~\cite{peng2023instruction} & - & general & 52 k & 21 & 163 \\
\bottomrule
\end{tabular}
\end{table*}

\subsection{Dataset and Evaluation Metric}
\label{app:dataset}
Table~\ref{tab:dataset} shows descriptions of these datasets, including information about the domain, evaluation metrics, number of samples, average length of instruction, and average length of response.

\paragraph{PubMedQA} PubMedQA\footnote{https://huggingface.co/datasets/axiong/pmc\_llama\_instructions}~\cite{jin2019pubmedqa} is a multiple-choice question-answering dataset optimized for medical reasoning. In this paper we utilize the version sourced from PMC-LLama \cite{wu2024pmc}. It features enhanced QA pairs with structured explanations derived from ChatGPT~\cite{roumeliotis2023chatgpt}, facilitating in-depth medical analysis. PubMedQA dataset consists of 211.3k training samples.

\paragraph{FiQA} FiQA dataset\footnote{https://huggingface.co/datasets/FinGPT/fingpt-fiqa\_qa} is a subset from FinGPT~\cite{yang2023fingpt}, which consists 17.1k financial open question-answers. We split out 200 samples for evaluation and adopted the MT-Bench instruction template (see Table~\ref{tab:mtbench}) to call ChatGPT~\cite{roumeliotis2023chatgpt} API (gpt-4-1106-preview). For the evaluation metric, we utilize the win rate to demonstrate the data quality ratio: $win\_rate=win\_counts/(win\_counts+lose\_counts)$.

\paragraph{AQUA\_RAT} 
The AQUA-RAT~\cite{aqua} dataset\footnote{https://huggingface.co/datasets/aqua\_rat} is a large-scale mathematical dataset with a collection of around 100k algebraic word problems. Each problem in the dataset is accompanied by a detailed, step-by-step solution narrative, articulated in natural language. 
This dataset consists of 97.5k training samples and 245 test samples. We use accuracy as the evaluation metric.

\paragraph{Mol-Instructions}
The Mol-Instructions~\cite{fang2023mol} dataset~\footnote{https://huggingface.co/datasets/zjunlp/Mol-Instructions}  consists of a subset: biomolecular text instructions, specifically designed for natural language processing tasks in bioinformatics and chemoinformatics. It encompasses six distinct information extraction and question-answering (Q\&A) tasks, structured through 53k detailed instructions. This design supports advanced NLP applications that require precise and context-specific understanding in the scientific domains of biology and chemistry. Our experiment only samples the open-Q\&A task with 37k training set and 1k test set. For evaluation, the BertSocre~\cite{zhang2019bertscore}, an automatic evaluation metric for text generation, is applied on a predefined test set of size 200.

\paragraph{Fed-WildChat} Fed-WildChat is a key component of the FedLLM-Bench~\cite{fedllm-bench}, a benchmark designed for evaluating FL methods in the context of LLMs. This dataset specifically focuses on multi-turn chat instruction tuning, providing a realistic representation of user-chatbot interactions. Fed-WildChat~\cite{zhao2024wildchat} is derived from a collection of conversations between humans and ChatGPT, WildChat, featuring a diverse array of interactions. It comprises data from 100 clients, totaling approximately 53,000 samples. This dataset is structured to reflect real-world scenarios by partitioning the data based on user IP addresses, ensuring that each client has a substantial number of samples (at least 200) for effective training and evaluation

\subsection{FedDQC algorithm}
To control the data quality for training, two steps need to be conducted: data selection with data quality assessment, and training process with high-quality data.
Since in FL, data is preserved at the client side, only the client could assess their data quality and select its data based on the data quality score.
In our FedDQC framework, data manipulations are mainly on the client side including the data quality measurement and local data training.
The key idea of this framework is to integrate data quality assessment with the training process, which consists of two components the alignment-based data quality assessment and the quality-aware hierarchical training. These components are detailed in Algorithm~\ref{alg:algorithm} and illustrated in Figure~\ref{fig:main}.

\begin{algorithm*}[t]
    \caption{FedDQC: Federated Data Quality Control}
    \label{alg:algorithm}
    \begin{algorithmic}[1] 
    \STATE \textbf{Initialization:} Initial global model: $\theta^0$; Training datasets: $\mathcal{D}=\{\mathcal{D}_1, \mathcal{D}_2, \dots, \mathcal{D}_N\}$; Number of training rounds: $R$; Number of hierarchies: $K$; Global quality threshold: $\lambda$

    \FOR{$k = 1$ to $K$}
        
        \STATE \textit{// Scoring Stage:} 
        \FOR{$n = 1$ to $N$}
            \STATE $\mathcal{D}_n = \mathcal{D}_n \setminus \mathcal{H}_{n(k-1)}$
            \COMMENT Remove the trained data from $\mathcal{D}_n$
            \STATE $\mathcal{S}_n=\{s_i: s_i=f_{IRA}((q^i,a^i)\in\mathcal{D}_n, \theta^{(R/K)*(k-1)})\}$
            \COMMENT Assess data quality of $\mathcal{D}_n$
        
            \STATE $\mathcal{D}'_n=\{(q^i, a^i)\in\mathcal{D}_n, s_i\geq\lambda\}$
            \COMMENT Select data points with quality scores above $\lambda$
            
        \ENDFOR
        
        \STATE \textit{// Training Stage:}
        \FOR{$r = (R/K)*(k-1)$ to $(R/K)*k-1$}
            \FOR{$n = 1$ to $N$}
                \STATE Sort $\mathcal{D}'_n$ by quality scores $s_i$ in descending order
                \STATE Split sorted $\mathcal{D}'_n$ into hierarchies $\mathcal{H}_{nk}, \dots, \mathcal{H}_{nK}$ with equal size $floor(|\mathcal{D}_n'|/(K-k)$
                \STATE \COMMENT Split local dataset to hierarchies
                \STATE Local update $\theta_n^r$ with $\mathcal{H}_{nk}$
                \COMMENT Local easy-to-hard hierarchical training

            \ENDFOR
            \STATE $\theta^{r+1}=\sum_{n=1}^N w_n \theta_n^{r,t}$
            \COMMENT Aggregate local models to update global model $\theta^r$
            \STATE Distribute global model $\theta^{r+1}$ to each client $n$
        \ENDFOR
        
    \ENDFOR
    
    \STATE \textbf{Return:} Global model $\theta^R$
    \end{algorithmic}    
\end{algorithm*}

\begin{table*}[t]
\centering
\caption{Comparison between the performance of high-quality data and low-quality data according to the IRA metric.}
\begin{tabular}{l|c|c|c|c}
\toprule
            & PubMedQA & AQUA-RAT & Mol-Instructions& FiQA \\
            & Acc      & Acc      & Acc          & Win rate \\
            \midrule
Full data & 0.750 & 0.2992 & 0.812 & - \\
High-score & 0.73 & 0.2559 & 0.822 & 0.7810 \\
Low-score  & 0.723 & 0.1732 & 0.800 & 0.3733 \\
\bottomrule
\end{tabular}
\label{tab:high_vs_low}
\end{table*}

\subsection{Baselines}
\label{app:baseline}
\paragraph{Comparisons with current methods} Compared to NUGGETS~\cite{nuggets} and AlpaGasus~\cite{chen2023alpagasus}, which utilize an external model for quality evaluation, FedDQC evaluates the data on the client side and preserves local data privacy. Unlike DataInf~\cite{kwon2023datainf} and NUGGETS~\cite{nuggets}, which require an extra validation set from the server, these methods become inapplicable in scenarios where the server cannot provide this set. Additionally, their computational cost is related to the size of the validation set. Compared to IFD~\cite{IFD}, FedDQC does not require extra dataset adaptation training, thus, is computation effective.

\paragraph{Perplexity}
Perplexity, a probability-based metric, is defined as the exponentiated average of the negative log-likelihoods of a tokenized sequence $X = (x_0, x_1, \ldots, x_t)$. Specifically, the perplexity of $X$, denoted as $\mathrm{PPL}(X)$, is calculated using the formula $\mathrm{PPL}(X) = \exp \left\{-\sum_i^t \log p_\theta(x_i \mid x_{<i})/t\right\}$, where $\log p_\theta(x_i \mid x_{<i})$ represents the log-likelihood of the $i^{th}$ token, conditional on its preceding tokens $x_{<i}$. This measure is frequently employed to data cleaning within a pre-trained corpus~\cite{wenzek2019ccnet}.

\paragraph{DataInf}
Influence functions, a gradient-based scoring method, rely on the model's performance on a validation set. DataInf, as introduced by~\cite{kwon2023datainf}, stands out as the first computationally efficient approximation of influence functions that can be practically implemented in LLMs. This Hessian-based standard influence functions, provide scores $\operatorname{DataInf}(x_j)_i = \nabla L(x_j; \theta^{\star}) H_{\theta^{\star}}^{-1} \nabla L(x_i; \theta^{\star})$ for every $x_i$ in $\mathcal{D}_k$ and $x_j$ in $\mathcal{D}_{val}$, where $\theta^{\star}$ denotes the parameters of the model trained on the training dataset, and $H_{\theta^{\star}}$ is the Hessian matrix of the empirical loss function. However, this method needs the model's convergence, which is unreal. To adapt to a federated setting, we first use the full dataset trained for 100 rounds for domain-specific datasets and 200 rounds for the general dataset. Then using this well-trained model to estimate the data influence score.

\paragraph{IFD} The Instruction-Following Difficulty (IFD) metric is calculated by the formula IFD\(_{\theta}(Q, A)=\frac{s_{\theta}(A|Q)}{s_{\theta}(A)}\), where \(s_{\theta}(A)=-\frac{1}{N}\sum_{i=1}^{N}logP(w_{i}^{A}|w_{1}^{A}, ..., w_{i-1}^{A}; \theta), s_{\theta}(A|Q)=-\frac{1}{N}\sum_{i=1}^{N}logP(w_{i}^{A}|Q,w_{1}^{A}, ..., w_{i-1}^{A}; \theta).\) IFD metric measures the difficulty of following instructions of a given sample. We train our model for 20 rounds on the targeted dataset, and subsequently, this pre-trained model is used for experiments with IFD as the scoring metric.

\paragraph{NUGGETS} NUGGETS leverages the disparity between one-shot and zero-shot scores to calculate a definitive gold score for each instruction. Exploiting the inherent contextual learning capabilities of large models.

\subsection{Experimental complements}
\subsubsection{Training setting}
\label{app:train_setting}
All the experiments are conducted on machines with the same hardware configuration using one NVIDIA GeForce RTX 4090. In all experiments, we use 8-bit quantization with batch size equal to 16, max length equal to 1024, and LoRA rank equal to 64 with a constant $\alpha=128$. For the federated setting, we consider $100$ communication rounds, $5$ clients with $8k$ training data in total for domain-specific dataset and $5$ clients with around $8k$ training data in total for Fed-WildChat dataset. We randomly sample $2$ clients for each round with $10$ local steps using AdamW~\cite{loshchilov2017decoupled} optimizer of model training. This setting is equivalent to 3 epochs for local training. For the NIID setting, we follow the Dirichlet distribution (with hyperparameter set to 5 for PubmedQA and FiQA, and 3 for AQUA-RAT and Mol-Instructions). We apply a cosine learning rate schedule according to the round index. The initial learning rate in the first round is $1e-4$, and the final learning rate in the last round is $1e-6$.  We use the Alpaca template~\cite{alpaca} to format the instruction, as shown in Appendix~\ref{app:tempalte}.

\subsubsection{How Data quality affects training performance}
We compare the high-score proportion of data with the low-score proportion of data and show that the data quality indeed affects training performance. See Table~\ref{tab:high_vs_low}.

\subsubsection{IRA metric analysis}
Here, simplicity and complexity refer to learning difficulty. We analyzed the correlation between IRA scores and gradient magnitudes, finding that higher IRA scores correspond to smaller gradients, which indicate easier learning.

\begin{figure}[h]
    \centering
    \vspace{-15pt}
    \includegraphics[width=0.8\linewidth]{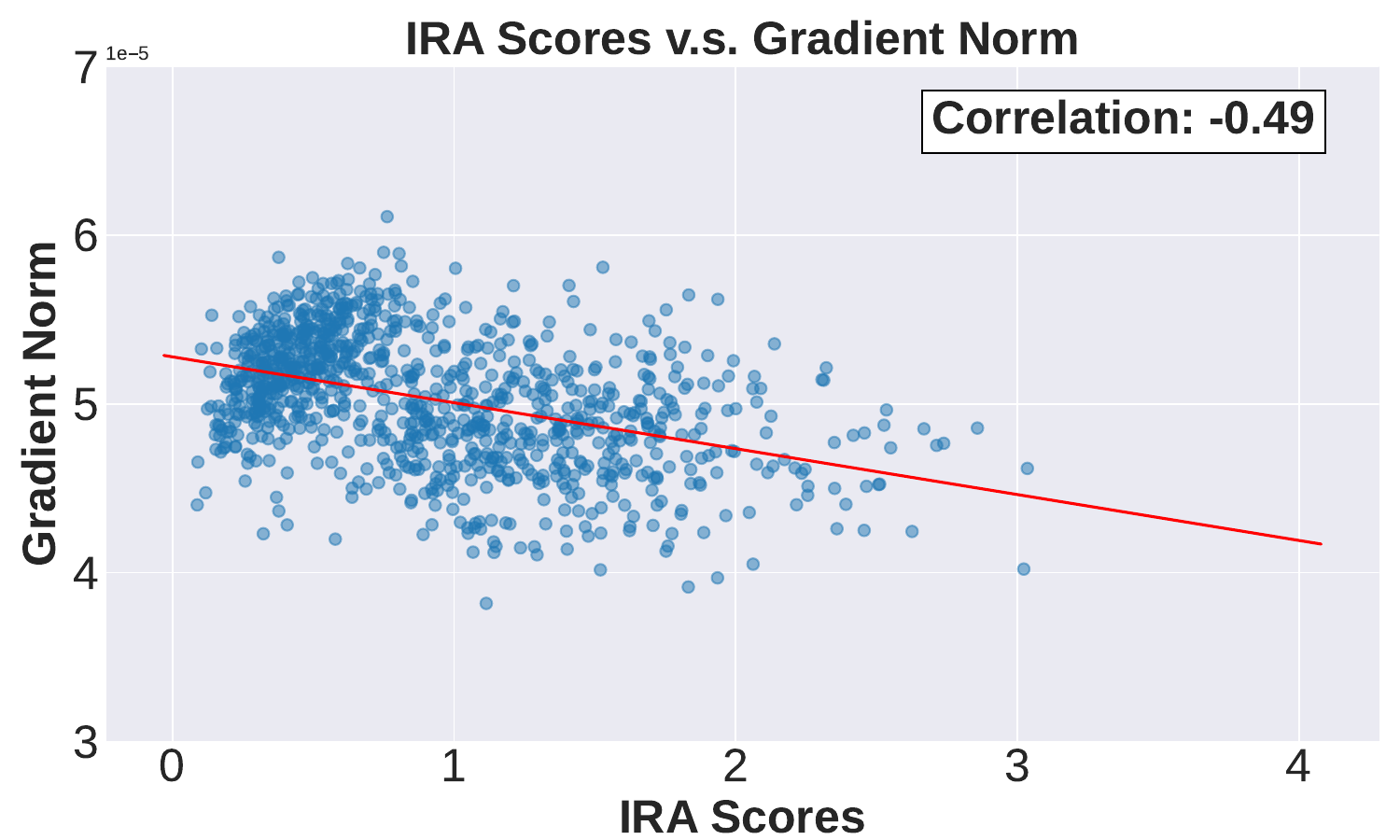}
    \caption{IRA score v.s. Gradient Norm}
    \vspace{-15pt}
    \label{fig:grad_IRA}
\end{figure}

\begin{table*}[t]
\caption{Comparison of the Proportion of High - Quality Samples Before and After Data Selection: An Analysis of the Performance of Different Methods}
\centering
\begin{tabular}{l|llllll}
\toprule
   & - & PPL & DataInf & IFD & NUGGESTS & IRA                  \\
\midrule
Data quality ratio & 0.5        & 0.8839       & 0.5003           & 0.700        & 0.6701            & \textbf{0.9345} \\
\bottomrule
\end{tabular}
\label{Tab:app_quality_ratio}
\end{table*}

\subsubsection{More FL settings.}
We compare more federated settings with the number of clients equal to 5 and 20. See Table \ref{tab:NIID}.

\begin{table*}[t]
\centering
\caption{Comparison with other NIID settings and client numbers.}
\begin{tabular}{l|cc|cccc}
\toprule
            & \multicolumn{2}{c|}{client = 20}   & \multicolumn{4}{c}{client = 5}                                                                      \\
            \hline
            & \multicolumn{1}{c|}{IID}   & IID   & \multicolumn{1}{c|}{NIID-0.1} & \multicolumn{1}{c|}{NIID-1} & \multicolumn{1}{c|}{NIID-5} & NIID-10 \\
            \hline
oracle      & \multicolumn{1}{c|}{0.741} & 0.750 & \multicolumn{1}{c|}{0.737}    & \multicolumn{1}{c|}{0.743}  & \multicolumn{1}{c|}{0.747}  & 0.758   \\
mix-quality & \multicolumn{1}{c|}{0.691} & 0.681 & \multicolumn{1}{c|}{0.662}    & \multicolumn{1}{c|}{0.655}  & \multicolumn{1}{c|}{0.664}  & 0.685   \\
selection   & \multicolumn{1}{c|}{0.743} & 0.751 & \multicolumn{1}{c|}{0.746}    & \multicolumn{1}{c|}{0.742}  & \multicolumn{1}{c|}{0.758}  & 0.751  \\ \bottomrule
\end{tabular}
\label{tab:NIID}
\end{table*}

\subsubsection{Other types of low-quality data}
We have supplemented the experiments by adding comparisons with other baselines under different bad data construction scenarios, as well as mixed types of bad datasets. All datasets have 50\% data corrupted.

\begin{table*}[t]
\caption{Performance comparison of FedAvg and various baseline methods under different bad data construction scenarios, with 50\% data corruption across different types of data alteration strategies. The best-performing results are highlighted in bold.}
\begin{tabular}{c|ccccccc|c}
\toprule
Bad type       & - & Swap  & Delete & Cut   & Substitute & Noisy & Mixture & Avg   \\
\midrule
FedAvg         & 0.757      & 0.681          & 0.691           & 0.730          & 0.720               & 0.734          & 0.700            & 0.709          \\
FedAvg+PPL     & -          & 0.703          & \textbf{0.722}  & 0.694          & 0.689               & 0.727          & 0.644            & 0.696          \\
FedAvg+DataInf & -          & 0.728          & 0.705           & 0.690          & 0.708               & 0.683          & 0.711            & 0.704          \\
FedAvg+IFD     & -          & 0.714          & 0.718           & 0.698          & 0.708               & 0.716          & 0.689            & 0.707          \\
FedAvg+NUGGETS & -          & 0.708          & 0.102           & 0.301          & 0.269               & 0.722          & 0.477            & 0.429          \\
\midrule
FedDQC         & 0.750      & \textbf{0.751} & 0.710           & \textbf{0.741} & \textbf{0.739}      & \textbf{0.737} & \textbf{0.731}   & \textbf{0.734} \\ \bottomrule
\end{tabular}
\end{table*}

The Table~\ref{Tab:app_quality_ratio} below shows the proportion of high-quality samples globally before and after data selection, referred to as the data quality ratio. A ratio closer to 1 indicates more high-quality data. Our method, IRA, maintains a higher proportion of high-quality data after selection.

\subsection{Prompt Template}
\label{app:tempalte}
\begin{table*}[h]
\caption{Alpaca Template for federated instruction tuning}
\label{tab:template_alpaca}
\begin{response}
Below is an instruction that describes a task. Write a response that appropriately completes the request.\\
\\
\#\#\# Instruction:\\
\{Instruction\}\\
\\
\#\#\# Response:
\end{response}
\end{table*}

\begin{table*}[h]
\caption{Alpaca Template for federated instruction tuning}
\label{tab:mtbench}

\begin{response}

[System]

Please act as an impartial judge and evaluate the quality of the responses provided by two AI assistants to the user question displayed below. You should choose the assistant that follows the user's instructions and answers the user's question better. Your evaluation should consider factors such as the helpfulness, relevance, accuracy, depth, creativity, and level of detail of their responses. Begin your evaluation by comparing the two responses and provide a short explanation. Avoid any position biases and ensure that the order in which the responses were presented does not influence your decision. Do not allow the length of the responses to influence your evaluation. Do not favor certain names of the assistants. Be as objective as possible. Don't provide your explanation, output your final verdict by strictly following this format: "[[A]]" if assistant A is better, "[[B]]" if assistant B is better, and "[[C]]" for a tie.

[User Question]

\{question\}

[The Start of Assistant A's Answer]

\{answer\_a\}

[The End of Assistant A's Answer]

[The Start of Assistant B's Answer]

\{answer\_b\}

[The End of Assistant B's Answer]
\end{response}

\end{table*}

\subsection{Case Study}
\label{app:case study}
\subsubsection{Examples of synthetic low-quality data}
The low-quality data we construct needs to be challenging for data cleansing and have a significant impact on performance. Therefore, we adopted a method of constructing low-quality data by swapping answers, simulating the scenario of incorrect data responses in real situations. Additionally, this construction method also maintains the content invariance of the corpus. Examples are presented below.

Table~\ref{tab:low-quality PubMedQA}, ~\ref{tab:low-quality FiQA}, ~\ref{tab:low-quality AQUA-RAT}, ~\ref{tab:low-quality Mol-Instruct} shows examples of synthetic low-quality data in 4 domain-specific datasets. Each dataset showcases a pair of data, including the questions, the authentic correct answers, and the incorrect answers generated by swapping the correct ones.  The synthetic low-quality data is created by pairing the same question with its incorrect answer.

\begin{table*}[t]
\centering
\caption{Synthetic low-quality data in PubMedQA}
\begin{tabularx}{\textwidth}{X|X|X} 
\toprule
\textbf{Question} & \textbf{Correct Answer} & \textbf{Incorrect Answer} \\
\midrule
Given your profession as a doctor, please provide responses to the medical questions using the patient's description.

Analyze the question given its context. Give both long answer and yes/no decision.

\#\#\#Question: Does mineralocorticoid receptor blockade inhibit accelerated atherosclerosis induced by a low sodium diet in apolipoprotein E-deficient mice?

\#\#\#Context: A low-sodium diet (LSD) was shown to increase both angiotensin II (AngII) and aldosterone levels ... & In E0 mice, Epl inhibited LSD-accelerated atherosclerosis, despite the elevation of renin and aldosterone levels. It is therefore suggested that the atherogenic action of LSD could be mediated, at least in part, by activation of the mineralocorticoid receptor. In addition, eplerenone may have direct anti-inflammatory actions.
\newline \newline
\#\#\#Answer: yes & 
Therefore, the technique could be considered an alternative method in increasing the keratinized tissue for cases that have a minimal amount of keratinized tissue.
\newline \newline
\#\#\#Answer: yes \\
\hline
In your capacity as a doctor, it is expected that you answer the medical questions relying on the patient's description.

Analyze the question given its context. Give both long answer and yes/no decision.

\#\#\#Question: Do simultaneous placement of an interpositional free gingival graft with nonsubmerged implant placement?

\#\#\#Context: Various surgical techniques target achieving adequate keratinized tissue around dental implants ...
& Therefore, the technique could be considered an alternative method in increasing the keratinized tissue for cases that have a minimal amount of keratinized tissue.
\newline \newline
\#\#\#Answer: yes
& In E0 mice, Epl inhibited LSD-accelerated atherosclerosis, despite the elevation of renin and aldosterone levels. It is therefore suggested that the atherogenic action of LSD could be mediated, at least in part, by activation of the mineralocorticoid receptor. In addition, eplerenone may have direct anti-inflammatory actions.
\newline \newline
\#\#\#Answer: yes \\

\bottomrule
\end{tabularx}
\label{tab:low-quality PubMedQA}
\end{table*}

\begin{table*}[t]
\centering
\caption{Synthetic low-quality data in FiQA}
\begin{tabularx}{\textwidth}{X|X|X} 
\toprule
\textbf{Question} & \textbf{Correct Answer} & \textbf{Incorrect Answer} \\
\midrule
Share your insights or perspective on the financial matter presented in the input.

\#\#\#Question: What one bit of financial advice do you wish you could've given yourself five years ago?",
& Compound interest. Next time you buy a 100\$ toy realize that if you save it - in x years that 100\$ you saved and invested could potentially be more than 100\$ where as most likely whatever you're buying will be worth much less. & If the building has no income, it also probably has minimal expenses. The heat, water and electricity costs are nearly zero. They are letting the value depreciate, and taking it off the taxes. I also suspect the condition of the building is poor, so any effort to make the building productive would be very costly. Many cities combat this by setting the tax on empty buildings or empty lots at a much higher rate. ...
\\
\hline
Utilize your financial knowledge, give your answer or opinion to the input question or subject . Answer format is not limited.

\#\#\#Question: Tax deductions on empty property
& If the building has no income, it also probably has minimal expenses. The heat, water and electricity costs are nearly zero. They are letting the value depreciate, and taking it off the taxes. I also suspect the condition of the building is poor, so any effort to make the building productive would be very costly. Many cities combat this by setting the tax on empty buildings or empty lots at a much higher rate. ...
& Compound interest. Next time you buy a 100\$ toy realize that if you save it - in x years that 100\$ you saved and invested could potentially be more than 100\$ where as most likely whatever you're buying will be worth much less. \\

\bottomrule
\end{tabularx}
\label{tab:low-quality FiQA}
\end{table*}

\begin{table*}[t]
\caption{Synthetic low-quality data in AQUA-RAT}
\centering
\begin{tabularx}{\textwidth}{X|X|X} 
\toprule
\textbf{Question} & \textbf{Correct Answer} & \textbf{Incorrect Answer} \\
\midrule
\#\#\#Instruction: Choose the correct option for the following math question.

\#\#\#Question: 1000 men have provisions for 15 days. If 200 more men join them, for how many days will the provisions last now?

\#\#\#Options:

A. 12.8

B. 12.4

C. 12.5

D. 16.8

E. 92.7

&
\#\#\#Rationale: 1000*15 = 1200*x

x = 12.5
\newline \newline
\#\#\#Answer: OPTION C IS CORRECT.
& \#\#\#Rationale: Explanation:

Let the sum of money be x

then

(x × 4 × 8)/100 = (560 × 12 × 8)/100

x × 4 × 8 = 560 × 12 × 8

x × 4 = 560 × 12

x = 560 × 3 = 1680
\newline \newline
\#\#\#Answer: OPTION D IS CORRECT.
\\
\hline
\#\#\#Instruction: Choose the correct option for the following math question.

\#\#\#Question: If simple interest on a certain sum of money for 8 years at 4\% per annum is same as the simple interest on Rs. 560 for 8 years at the rate of 12\% per annum then the sum of money is

\#\#\#Options:

A. Rs.1820

B. Rs.1040

C. Rs.1120

D. Rs.1680

E. None of these &
\#\#\#Rationale: Explanation:

Let the sum of money be x

then

(x × 4 × 8)/100 = (560 × 12 × 8)/100

x × 4 × 8 = 560 × 12 × 8

x × 4 = 560 × 12

x = 560 × 3 = 1680 
\newline \newline
\#\#\#Answer: OPTION D IS CORRECT.
&
\#\#\#Rationale: 1000*15 = 1200*x

x = 12.5
\newline \newline
\#\#\#Answer: OPTION C IS CORRECT.

\\
\bottomrule
\end{tabularx}
\label{tab:low-quality AQUA-RAT}
\end{table*}

\begin{table}[t]
\centering
\caption{Synthetic low-quality data in Mol-Instructions}
\begin{tabularx}{\textwidth}{X|X|X} 
\toprule
\textbf{Question} & \textbf{Correct Answer} & \textbf{Incorrect Answer} \\
\midrule
Answer this question truthfully

\#\#\#Question: What is the predicted relative molecular mass of the protein encoded by PVAS2?
& The predicted relative molecular mass of the protein encoded by PVAS2 is 65810 Da.
& Resonance Raman spectroscopy is a form of spectroscopy used to analyze the vibrational, rotational, and other structural characteristics of molecules.
\\
\hline
Answer this question truthfully

\#\#\#Question: What is Resonance Raman spectroscopy?
& Resonance Raman spectroscopy is a form of spectroscopy used to analyze the vibrational, rotational, and other structural characteristics of molecules.
& The predicted relative molecular mass of the protein encoded by PVAS2 is 65810 Da. \\

\bottomrule
\end{tabularx}
\label{tab:low-quality Mol-Instruct}
\end{table}

\subsubsection{Examples of scored data}
Figures ~\ref{score data: PubMedQA}, ~\ref{score data: FiQA}, ~\ref{score data: AQUA_RAT}, ~\ref{score data: Mol-Instructions} shows examples of scored data in 4 mixed-quality domain-specific datasets. Each dataset's size is 8k, with 50\% low-quality data generated by swapping correct answers. The remaining 50\% is considered high-quality data. We use IRA as the scoring metric and show typical data examples with scores in top 1\% and lowest 1\%.\\ \\
Typically, high-quality data scores high and low-quality data scores low. This is because the incorrect answers in low-quality data significantly diminish the instruction-response relativeness, leading to an increase in IRA. However, the high-quality data example in Figure ~\ref{score data: PubMedQA} scores low, due to the presence of complicated and verbose input. Consequently, the model finds it challenging to establish the relativeness between the instruction and response.

\begin{figure*}[!t]
  \centering
  \parbox{\textwidth}{
        \rule{\textwidth}{1.5pt} 
        \centering 
        \textbf{Scored data examples in PubMedQA} 
        \rule{\textwidth}{0.8pt} 

        \begin{minipage}[t]{0.47\textwidth}
            \textcolor{teal}{High-quality, High-score}\\
            \textcolor{teal}{IRA score: 4.08} \\ \\
            \textcolor{teal}{[Instruction]} \\Considering your role as a medical practitioner, please use the patient's description to answer the medical questions.\\
            Analyze the question given its context. Give both long answer and yes/no decision.
            
            \textcolor{teal}{[Input]} \\\#\#\#Question: Does [ Hemorrhagic shock increase the occurrence of bacterial translocation ]?
            
            \#\#\#Context: To determine whether hemorrhagic shock (HS) increases the occurrence of bacterial translocation (BT). 100 patients were divided into 4 groups: control group (group I, 34 patients); group with hemorrhagic shock (HS) caused by closed blunt abdominal trauma (group II, 23); group caused by closed blunt abdominal trauma without HS (group III 15); and group with HS caused by intra-abdominal viscus hemorrhage (group IV 28). Preoperative and postoperative samples were taken from peripheral blood, visceral peritoneal swab, portal vein blood, ileal mesenteric lymph node, liver and spleen biopsy respectively for aerobic and anaerobic culture. The positive culture rates of these groups were 6\%, 65\%, 13\%, 68\% respectively. The difference between the control and experimental groups was significant(P < 0.05). The difference was also significant between group I and II and between I and IV (P < 0.01), whereas it was not significant between I and III, and between II and IV (P > 0.05).
            
            \textcolor{teal}{[Output]}\\
            HS increases the occurrence of BT.\\ \\\#\#\#Answer: yes

            
        \end{minipage}
        \hfill
        \begin{minipage}[t]{0.47\textwidth}
            \textcolor{teal}{High-quality, Low-score}\\
            \textcolor{teal}{IRA score: -0.61} \\ \\
            \textcolor{teal}{[Instruction]}\\
            Given your background as a doctor, please provide your insight in addressing the medical questions based on the patient's account.\\Analyze the question given its context. Give both long answer and yes/no decision.
            
            \textcolor{teal}{[Input]}\\\#\#\#Question: Does globulin-platelet model predict minimal fibrosis and cirrhosis in chronic hepatitis B virus infected patients?\\\#\#\#Context: To establish a simple model consisting of the routine laboratory variables to predict both minimal fibrosis and cirrhosis in chronic hepatitis B virus (HBV)-infected patients. We retrospectively investigated 114 chronic HBV-infected patients who underwent liver biopsy in two different hospitals. Thirteen parameters were analyzed by step-wise regression analysis and correlation analysis. A new fibrosis index [globulin/platelet (GP) model] was developed, including globulin (GLOB) and platelet count (PLT). GP model = GLOB (g/mL) × 100/PLT (× 10(9)/L). We evaluated the receiver operating characteristics analysis used to predict minimal fibrosis and compared six other available models. Thirteen clinical biochemical and hematological variables [sex, age, PLT, alanine aminotransferase, aspartate aminotransferase (AST), albumin, GLOB, total bilirubin (T.bil), direct bilirubin (D.bil), glutamyltransferase, alkaline phosphatase, HBV DNA and prothrombin time (PT)] were analyzed according to three stages of liver fibrosis (F0-F1, F2-F3 and F4). Bivariate Spearman's rank correlation analysis showed that six variables, including age, PLT, T.bil, D.bil, GLOB and PT, were correlated with the three fibrosis stages (FS). Correlation coefficients were ... 
            
            \textcolor{teal}{[Output]} \\
            By establishing a simple model using available laboratory variables, chronic HBV-infected patients with minimal fibrosis and cirrhosis can be diagnosed accurately, and the clinical application of this model may reduce the need for liver biopsy in HBV-infected patients.\\ \\\#\#\#Answer: yes 
        \end{minipage}

        \rule{\textwidth}{0.8pt} 

  }
\caption{
The examples of scored data in mixed-quality PubMedQA dataset include a high-quality datapoint with a high score and a high-quality datapoint with a low score.
} 
\label{score data: PubMedQA} 
\end{figure*}

\begin{figure*}[!t]
  \centering
  \parbox{\textwidth}{
        \rule{\textwidth}{1.5pt} 
        \centering 
        \textbf{Scored data examples in FiQA} 
        \rule{\textwidth}{0.8pt} 

        \begin{minipage}[t]{0.47\textwidth}
            \textcolor{teal}{High-quality, High-score}\\ 
            \textcolor{teal}{IRA score: 3.16} \\ \\
            \textcolor{teal}{[Instruction]} \\Share your insights or perspective on the financial matter presented in the input.\\
            \textcolor{teal}{[Input]} \\\#\#\#Question: On a debt collection agency's letter, what does “balance” refer to?
            
            \textcolor{teal}{[Output]}\\
            The balance is the amount due.
            
        \end{minipage}
        \hfill
        \begin{minipage}[t]{0.47\textwidth}
            \textcolor{teal} {Low-quality, Low-score}\\
            \textcolor{teal}{IRA score: -0.28} \\ \\
            \textcolor{teal}{[Instruction]} \\Utilize your financial knowledge, give your answer or opinion to the input question or subject . Answer format is not limited.\\
            \textcolor{teal}{[Input]} \\\#\#\#Question: How can a person with really bad credit history rent decent housing?
            
            \textcolor{teal}{[Output]}\\
            I can't give you a detailed answer because I'm away from the computer where I use kMyMoney, but IIRC to add investments you have to create new transactions on the 'brokerage account' linked to your investment account.  

        \end{minipage}

        \rule{\textwidth}{0.8pt} 

  }
\caption{
The examples of scored data in mixed-quality FiQA datasets include a high-quality datapoint with a high score and a low-quality datapoint with a low score.
} 
\label{score data: FiQA} 
\end{figure*}

\begin{figure*}[!t]
  \centering
  \parbox{\textwidth}{
        \rule{\textwidth}{1.5pt} 
        \centering 
        \textbf{Scored data examples in AQUA-RAT} 
        \rule{\textwidth}{0.8pt} 

        \begin{minipage}[t]{0.47\textwidth}
            \textcolor{teal} {High-quality, High-score}\\ 
            \textcolor{teal}{IRA score: 1.92} \\ \\
            \textcolor{teal}{[Instruction]} \\\#\#\#Instruction: Choose the correct option for the following math question.\\
            \textcolor{teal}{[Input]} \\\#\#\#Question: What should come in place of question mark(?) in the following equation? \\
            8597 – ? = 7429 – 4358\\\#\#\#Options:\\A. 5706\\B. 5526\\C. 5426\\D. 5626\\E. None of these\\
            
            \textcolor{teal}{[Output]}\\
            \#\#\#Rationale: 8597 – ? = 7429 – 4358\\  ? = 5526\\ \\\#\#\#Answer: OPTION B IS CORRECT.
            
        \end{minipage}
        \hfill
        \begin{minipage}[t]{0.47\textwidth}
            \textcolor{teal} {Low-quality, Low-score}\\
            \textcolor{teal}{IRA score: -0.04} \\ \\
            \textcolor{teal}{[Instruction]} \\\#\#\#Instruction: Choose the correct option for the following math question.\\
            \textcolor{teal}{[Input]} \\\#\#\#Question: A jar contains only red, yellow, and orange marbles. If there are 3 red, 5 yellow, and 4 orange marbles, and 2 marbles are chosen from the jar at random without replacing any of them, what is the probability that 2 yellow, 1 red, and no orange marbles will be chosen?\\\#\#\#Options:\\A. 1/60\\B. 1/45\\C. 2/45\\D. 3/22\\E. 6/22\\
            
            \textcolor{teal}{[Output]}\\
            \#\#\#Rationale: P= 16 = 16+3 = 19 = S\\O = 15 =15+3 =18 = R\\SIMILARLY,\\P = 16 = 16+3 = 19 = S\\R = 18 = 18+3 = 21 = U\\I = 9 = 9+3 = 12 = L\\Y = 25 = 25 +3 = 28 = B\\A = 1 = 1+3 = 4 = D\\SO ANS IS\\SULB\\ \\\#\#\#Answer: OPTION D IS CORRECT.  

        \end{minipage}

        \rule{\textwidth}{0.8pt} 

    }
\caption{
The examples of scored data in mixed-quality AQUA-RAT dataset include a high-quality datapoint with a high score and a low-quality datapoint with a low score.
} 
\label{score data: AQUA_RAT} 
\end{figure*}

\begin{figure*}[!t]
  \centering
  \parbox{\textwidth}{
        \rule{\textwidth}{1.5pt} 
        \centering 
        \textbf{Scored data examples in Mol-Instructions} 
        \rule{\textwidth}{0.8pt} 

        \begin{minipage}[t]{0.47\textwidth}
            \textcolor{teal} {High-quality, High-score}\\ 
            \textcolor{teal}{IRA score: 3.95} \\ \\
            \textcolor{teal}{[Instruction]} \\Answer this question truthfully\\
            \textcolor{teal}{[Input]} \\\#\#\#Question: How long after the beginning of the secondary aggregation process does the mitotic coefficient reach a first maximum?
            \textcolor{teal}{[Output]}\\The mitotic coefficient reaches a first maximum 8 hours after the beginning of the secondary aggregation process.\\
            
        \end{minipage}
        \hfill
        \begin{minipage}[t]{0.47\textwidth}
            \textcolor{teal} {Low-quality, Low-score}\\ 
            \textcolor{teal}{IRA score: -0.63} \\ \\
            \textcolor{teal}{[Instruction]} \\Answer this question truthfully\\
            \textcolor{teal}{[Input]} \\\#\#\#Question: What are system invariants?
            \textcolor{teal}{[Output]}\\The results suggest that multiple pathways exist in soybean to regulate expression of Aox genes.\\

        \end{minipage}

        \rule{\textwidth}{0.8pt} 

  }
\caption{
The examples of scored data in mixed-quality Mol-Instructions dataset include a high-quality datapoint with a high score and a low-quality datapoint with a low score.
} 
\label{score data: Mol-Instructions} 
\end{figure*} 

\end{document}